\documentclass[a4paper,12pt]{report}

\usepackage[utf8x]{inputenc}
\usepackage[english]{babel}
\usepackage[T1]{fontenc}
\usepackage{lmodern}

\usepackage{amsmath}
\usepackage{amssymb}
\usepackage{geometry}
\usepackage{a4wide}
\usepackage{enumerate}
\usepackage{graphicx}
\usepackage{lastpage}
\usepackage{cite}
\usepackage[pdftex]{hyperref}
\usepackage{caption}
\usepackage{subcaption}

\usepackage{fancyhdr}
\newcommand{\myname}{Axel Angel}
\newcommand{\mytitle}{Towards Distortion-Predictable Embedding of Neural Networks}
\newcommand{\mysubtitle}{Thesis}
\newcommand{\mydate}{June 18, 2015}

\newcommand{\N}{\mathbb{N}}

\newcommand{\R}{\mathbb{R}}

\date{\mydate}

\makeatletter
\AtBeginDocument{
  \hypersetup{
    pdfauthor = {\myname},
    pdftitle = {\mytitle},
    pdfsubject = {\mysubtitle},
    pdfcreationdate = {D:20150618000000},
  }
}
\makeatother

\newcommand{\eg}{e.g.}
\setkeys{Gin}{width=0.75\textwidth}

\begin{document}
\begin{titlepage}
\begin{center}
    \vspace*{0.5cm}

    \line(1,0){400}
    \vspace{0.5cm}

    {\bf \Large Towards Distortion-Predictable Embedding}

    \vspace{0.5cm}

    {\bf \Large of Neural Networks}

    \vspace{0.5cm}
    \line(1,0){400}

    \vspace{1.5cm}

    {\bf \myname}

    Master of Science

    School of Computer Science

    \vspace{1.6cm}

    Supervised by

    {\bf Prof. Pascal Fua}

    {\bf Sironi Amos}

    \vspace{0.8cm}

    Computer Vision Laboratory (CVLAB)

    École Polytechnique Fédérale de Lausanne (EPFL)

    Switzerland

    \vspace{3cm}

    \mydate

    \vfill

    \includegraphics[width=0.3\textwidth]{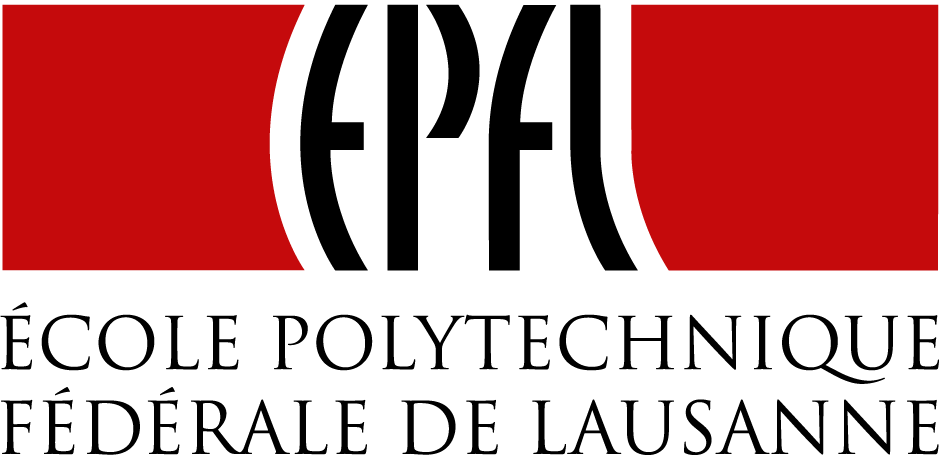}

    \end{center}
\end{titlepage}

\newpage
\thispagestyle{empty}
\vfill
\begin{center}
    Page intentionally left blank.
\end{center}

\cleardoublepage
\begin{abstract}
    Current research in Computer Vision has shown that {\em Convolutional Neural Networks} (CNN) give state-of-the-art performance in many classification tasks and Computer Vision problems\cite{mnist_web}\cite{krizhevsky2012imagenet}\cite{rowley1998neural}\cite{prechelt1994proben1}.
    The embedding of CNN, which is the internal representation produced by the last layer, can indirectly learn topological and relational properties.
    Moreover, by using a suitable loss function, CNN models can learn invariance to a wide range of non-linear distortions such as rotation, viewpoint angle or lighting condition.
    In this work, new insights are discovered about CNN embeddings and a new loss function is proposed, derived from the contrastive loss, that creates models with more predicable mappings and also quantifies distortions.
    In typical distortion-dependent methods, there is no simple relation between the features corresponding to one image and the features of this image distorted.
    Therefore, these methods require to feed-forward inputs under every distortions in order to find the corresponding features representations.
    Our contribution makes a step towards embeddings where features of distorted inputs are related and can be derived from each others by the intensity of the distortion.
\end{abstract}

\tableofcontents

\chapter{Introduction}

Nowadays, Computer Science has become predominant in many fields of science.
Data storage, data analysis and visualization are only a few examples where it plays a fundamental role.
The amount of available data is increasing every day which encourages the improvement of computers and the exploration of more complex problems.
Therefore, there is an enormous flow of information in terms of quantity and dimensionality.
Sound, images and videos are common example of large multidimensional data.
Many areas of science need methods to reliably extract specific information from these growing amount of data.
This is usually done through analysis, visualization or a combination of both.

Computer Vision (CV) is a field addressing the problem of analyzing visual multimedias by automatic processing.
Image classification and pattern recognition are two examples of problems studied in Computer Vision.
Many state-of-the-art solutions for these problems are combining the advance of Machine Learning to find data-driven models, instead of hand-designing algorithms, which usually requires expert knowledge of the domain.
Among these methods, {\em Convolutional Neural Networks} (CNN), are a variant of {\em Neural Network}s (NN) used in vision tasks and are now widely employed for Computer Vision with great success\cite{krizhevsky2012imagenet}\cite{rowley1998neural}\cite{prechelt1994proben1}.
They are trained in a supervised way to perform a systematic processing task, for instance to classify images into a number of fixed categories (\eg: handwritten digits between 0 and 9).
Their {\em features} are the internal representation of the input inside NN that is being optimized during training to give better results.
In common classification tasks, the features are meant to lie in a low dimensional space and should provide a more compact representation of the input.
Therefore, these NN can be seen as two embedded parts: a dimensionality reduction part that extracts important informations and a classifier that takes a decision based on these features.
In practical applications, image distortions of various kind are present in the data: shearing, noise, camera viewpoint, spatial positioning and lighting condition.
In the case of classification, distortions can require putting extra efforts to train a robust model able to classify correctly the input in the presence of such deformations.
Most implementations decide to nullify the distortion signal with data-augmentation which assigns the original label on distorted samples.
This can be done when the application does not require this information.
However, there are practical cases, for instance in Biomedical Imaging and Face Detection, which may require to have a feature representation able to quantify and possibly predict the distortions.

The first step of this work is to better understand NN embeddings and explore if conventional tools can be used to extract distortion features reliably.
Moreover, visual inspection of the feature space using a human-friendly representation can help to gain insights and to improve models.
One way to analyze the features is to apply dimensionality reduction techniques to project highly-dimensional points into lower dimensional space.
Many differences characterize dimensionality reduction methods such as: properties, flexibility and final goals.
A brief overview is describing: PCA, MDS, LLE, Isomap and the more recent SNE.
However many of them lack flexibility by representing only linearity, while the others don't preserve hierarchical structures accurately\cite{t-SNE}.
Researches have shown that t-SNE, a variant of SNE, has greatly improved on this regard: it is capable of keeping global clusters and fine-grained coherence at small scales.
Moreover it was used multiple times successfully to visualize the feature space of NN\cite{donahue2013decaf}\cite{yu2014visualizing}.

Nonetheless, several limitations of t-SNE are encountered in this work.
It cannot be controlled directly because t-SNE is only given unlabelled points like unsupervised methods.
Therefore, as shown later, the resulting embedding can have an unexpected shape unfit for the target application.
Secondly, this method works directly on the representations of the points without creating a mapping that computes the relation between inputs and outputs.
Thus it is impossible to compute the representation of points that were not part of the optimization process at the beginning.
It would be necessary to recompute the whole representation from scratch with these new points but t-SNE is expensive.
In this work, a first try is made by combining CNN and t-SNE to produce an embedding quantifying distortions.
However, the results are not predictable and difficult to interpret.
Distortions are the primary factor of clustering, while classes are not well separated.
Moreover, problems were faced to scale this method on larger datasets.
t-SNE could be probably improved to leverage external prior-informations such as labels, but because of the major shortcomings faced, we opted for another solution.

An alternative dimensionality reduction method consists of using the NN to directly compute an embedding similar to t-SNE but trained in a supervised manner.
The network is used to learn a mapping from the image space to a lower embedding which does not have the previous shortcomings.
As NN have the capacity to represent any non-linear continuous functions\cite{csaji2001approximation}, it is suitable to compute a very powerful dimensionality reduction method.

To achieve this goal, {\em Dimensionality Reduction by Learning an Invariant Mapping} ({\em DrLIM}) combines concepts similar to the ones employed in t-SNE but expressed in an equivalent adapted formulation for NN\cite{hadsell2006dimensionality}.
The key elements are: the contrastive loss function with the Siamese network architecture for training and a pairing strategy directly describing the embedding.
The current formulation allows to express a one-dimensional relation between two pair of images based on the similarity: similar pairs of images should be close together whereas dissimilar pairs should be far from each other.
In this work, an extension of DrLIM is proposed with a generalized loss function.
It allows to express richer similarity relations between image pairs for training.
Moreover, each of these similarities is expressed in the embedding by dimensions selected in advance and is associated feature components.

Therefore, thanks to this extension, it becomes possible to have a better control on the final embedding of a NN.
The experiments with DrLIM are reproduced on the data-augmentation of the MNIST dataset\cite{lecun1998mnist} with translations or rotations and likewise on the NORB dataset\cite{lecun2004learning}.
The resulting embeddings are studied in details with multiple comparisons to the original work on DrLIM.
We find it is possible to embed distortion information in one extra dimension while preserving the characteristics of DrLIM on MNIST.
We also find that this new solution can represent the DrLIM's coherent and cyclic space of the camera viewpoint on NORB.
The successful application on these two datasets demonstrates the effectiveness of this method.

\section{Thesis Outline}
In Chapter~\ref{chap:related}, the previous papers related to this work are presented.
An overview of the prior art is describing dimensionality reduction methods and the advance of Computer Vision with NN.
Then, the reference paper introducing DrLIM models is discussed.

In Chapter~\ref{chap:neural_network}, a summary is given for the background knowledge necessary to understand the networks used in this work.
The general architecture of NN, their similarity with CNN and some important definitions are also discussed.

This is followed by Chapter~\ref{chap:dim_red} where the standard dimensionality reduction methods are reviewed in more details.
The major advantages and drawbacks of t-SNE, compared to other methods are explained.
Later in the chapter, the basic theoretical knowledge behind t-SNE optimisation problem is provided.
Then the necessary tools are explained to achieve dimensionality reduction with NN and how to extend the contrastive loss to N-dimensional similarity.

In Chapter~\ref{chap:results}, the experiment environment is established in terms of technical choices (software, network architecture and datasets).
This chapter also includes all the practical results, the hypothesises and also the early conclusions of our work.

In Chapter~\ref{chap:conclusion}, some final remarks are made about the results and we encourage to continue from our contribution by providing some propositions of future works.

\chapter{Related Work}
\label{chap:related}

More than 20 years ago, researchers discovered that NN is a highly flexible model architecture to solve many classification problems.
Most of today's NN are using convolutions in CNN which are a powerful and specialized variant for visual tasks like MNIST \cite{mnist_web} and ImageNet \cite{krizhevsky2012imagenet}, face detection as well \cite{rowley1998neural} and many more \cite{prechelt1994proben1}.
This architecture is more efficient for Computer Vision problems, partly because the convolutions share their weights in the first layers to extract spatial cues.

Current researches are mainly focused on advancing into more complex classification problems using Deep Learning.
This field suggests improving the current results by deepening the architecture in terms of number of layers to express more abstract and higher-level concepts.
Unfortunately, more layers require more computations than before due to the increase of parameters but researchers started to overcome this challenge.
Ways to speed up the training and the classification appeared and were greatly beneficial for the development of this field in recent years\cite{ciresan2011flexible}\cite{schmidhuber2015deep}\cite{nasse2009face}.
The use of GPUs parallelism and cloud computing allowed scaling up to much deeper architecture (with many more parameters) than before\cite{coates2013deep}.
Both researchers and professional programmers built various frameworks to train and to use NN with many different goals such as: performance, accessibility or composability.
The most populars are: Theano\cite{bastien2012theano}, Caffe\cite{jia2014caffe}, Torch7\cite{collobert2011torch7}, Pylearn2\cite{goodfellow2013pylearn2}.
The efficiency and modularity proposed by Caffe were leveraged in this work for several reasons described later.

Even though CNN models are now widely used in Computer Vision with great success thanks to these frameworks, current research exposed the ignorance of these networks behaviors.
Some papers work on reliable ways to fool networks using adverserial attacks which exploits their unintuitive properties\cite{szegedy2013intriguing}.
Most publications do not try to formally justify their good results because of the non-linear and complex relations between units. 
Moreover, this problem is greater with the addition of distortions in the inputs.
It is not possible to directly look at the high-dimensional embedding, because humans cannot easily plot nor interpret so many dimensions.
The manifestation of this relation can be looked visually using methods to produce an alternative human-friendly representation.

Dimensionality reduction methods is a popular tool in such a case\cite{dai2014document}\cite{taylor2011learning} and several methods exist that are specialized for visualization.
Standard methods, like PCA and MDS\cite{cox2000multidimensional}, are used for reducing dimensions usually as a preprocessing step and they are limited to linearity in the data.
Their primary characteristic is to maximize the variance of the original data which is important for reconstruction but not necessarily for visualization.
Other non-linear dimensionality reduction methods, for instance: Isomap\cite{tenenbaum2000global}, LLE\cite{roweis2000nonlinear} and SNE\cite{SNE}, solve an optimisation problem with a loss function which improve their flexibility for purposes like visualization and preserve local structure/cluster.
However, recent papers show that SNE has better potential to keep global clusters and local details at the same time unlike the other previously cited methods\cite{SNE}.

SNE has introduced a better maximization problem to preserve point neighborhood and general clusters with an important emphasis on distances\cite{SNE}.
However it suffers from ``center-crowdedness'' and faces difficulties in optimisation\cite{t-SNE}.
This is the reason why recent works are now using t-SNE, a variant of SNE, to reduce CNN embeddings into a 2D human-friendly manifold. 
This method exhibits an easier optimisation formulation and preserves more structures at diverse scales.
More convincing examples were created with t-SNE directly applied on popular datasets\cite{van2009new} like MNIST\cite{t-SNE}.
A more recent work proposed an alternative method, called Barnes-Hut SNE, that approximates t-SNE with a much faster algorithm but it wasn't useful for this work\cite{van2013barnes}.

Many more details in CNN embeddings emerge by using t-SNE.
For instance, how samples are grouped depending on multiple factors: essentially based on similarity of the digits (strokes, thickness and shapes) and natural variance (rotation).
Many papers use dimensionality reduction to create a human viewable representation of the output embedding \cite{donahue2013decaf}\cite{yu2014visualizing}\cite{yaotiny}.
Only a few papers actually try to formalize the input-output relations\cite{goodfellow2009measuring}.
Some papers proposes to learn transformation-invariant embeddings, by means of modelling distortions directly into their models\cite{gens2014deep} or using data-augmentation\cite{hadsell2006dimensionality}.

The latter proposes to create a dimensionality reducing mapping, called {\em Dimensionality Reduction by Learning an Invariant Mapping} ({\em DrLIM}), by training a CNN using a new loss function inspired by Energy-Based modeling, called the {\em contrastive loss}.
This paper will be mentioned in this work as the {\em reference paper} or DrLIM's paper.
Their results on the MINST dataset\cite{lecun1998mnist} shows that CNN can learn a mapping that distinguish labels, and group alike digits even when they are translated artificially.
They experimented with the NORB dataset\cite{lecun2004learning} as well with intriguing results: images are mapped on a 3D cylinder whose axes quantify the orientation in 2D and the azimuth angle in 1D.
The network was successfully trained to ignore lighting conditions, which is a strongly non-linear distortion.
They use the Siamese training architecture to present image pairs which are optimized to be close if similar or far otherwise\cite{bromley1993signature}\cite{chopra2005learning}.
There are several advantages of such a method: the cost penalty is very low and present only during training.
The usage of standard CNN also allows more freedom for experimenting with different architectures and the proposed loss function is effective while simple.

However the input-output relations of distorted images is usually damped instead of being quantified in a predictable way.
The distortion information should be kept because they can be valuable later on.
Besides, the reference paper provides subjective comments of the embedding coherence (descriptions and figures) but they do not offer nor propose a formal way to measure and compare its quality objectively.

In this work, a few steps are presented towards predictable embedding with respect to distortions and a simple qualitative measure is presented to compare similar methods.

\chapter{Theory: Neural Networks}
\label{chap:neural_network}
The general layout of neural networks and their mathematical foundations is now introduced.
The important differences between {\em Neural Networks} (NN) and {\em Convolutional Neural Networks} (CNN) are discussed and a justification is given why CNN is so predominant in Computer Vision tasks.
An intuitive explanation of embeddings is given and will be used later for dimensionality reduction.

\section{NN and CNN Classifiers}

\begin{figure}[t]
    \begin{center}
        \includegraphics{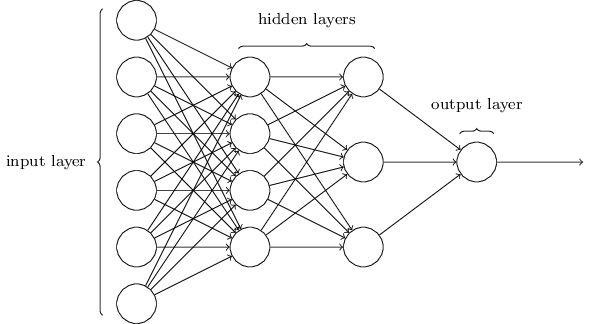}
    \end{center}
    \caption{An example of a Neural Network with three layers -- Source: neuralnetworksanddeeplearning.com}
    \label{fig:neural_network}
\end{figure}

Let us quickly recall the definition of a Neural Network and its workings for classification.
Basically, NN are composed of $L$ layers which are made of inter-connected neurons (see figure~\ref{fig:neural_network} for an example).
A connection $l_{i \rightarrow j}$ in the $k$th layer is weighted by a parameter $w^k_{i,j}$ which is learned.
Classification is done by feed-forwarding the input onto the first layer whose output is propagated layers by layers through the network.
The output of one layer is forwarded onto the next through their neuronal connections.
This process is repeated until the last layer, whose output is considered as the network output.
Each layer does a single particular computation over its input.
Usual NN have a repetition of the following pattern: one layer computing inner-products with their neuronal weights then adding biases, followed by a non-linear layer using an activation function.

\begin{figure}[t]
    \begin{center}
        \includegraphics{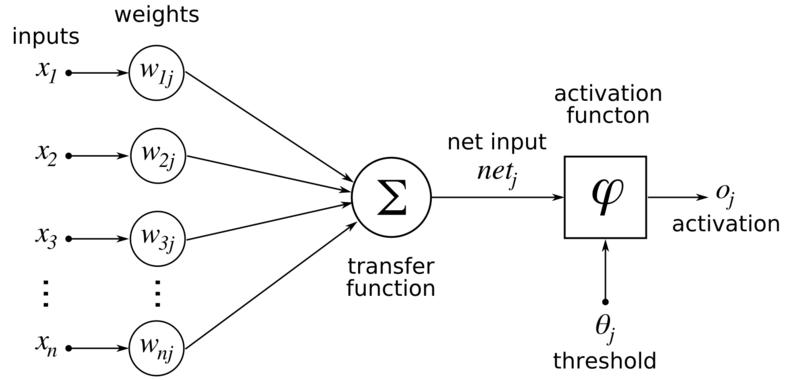}
    \end{center}
    \caption{A neuron computing an inner-product with a link to its activation neuron -- Source: Wikipedia}
    \label{fig:artificial_neurons}
\end{figure}

The figure~\ref{fig:artificial_neurons} presents the computation involving a particular neuron and its activation.
Let us define the $j$th neuron in the $k$th layer for the inner-product case, then its output $z^k_j$ is defined:
\begin{eqnarray}
    z^k_j = \sum_{i=1}^{U^{k-1}} w^k_{i,j} z^{k-1}_i + b^k_j
\end{eqnarray}
where $U^k$ is the number of neurons in layer $k$.
Then we define the $j$th neuron in the $k$th layer for the activation case:
\begin{eqnarray}
    z^k_j = g(z^{k-1}_j, w^k_{j})
\end{eqnarray}
where $g$ is the activation function with its parameter.
Most of them use the sigmoid activation function and connect each neuron to all the next layer neurons.
Training the weights generally involves gradient descent to minimize the loss function of such networks.
This loss represents the error between the prediction of the network (its output) versus the expected output (in classification: the class label).
The training involves an iterative process where: the input is first feed-forwarded into the network to compute the error, then this latter is back-propagated up to the first layer, while each unit weights is adjusted to minimize the unit error.
This process should be repeated until the error on the validation set has converged to a local optimum.

\begin{figure}[t]
    \begin{center}
        \includegraphics[width=\textwidth]{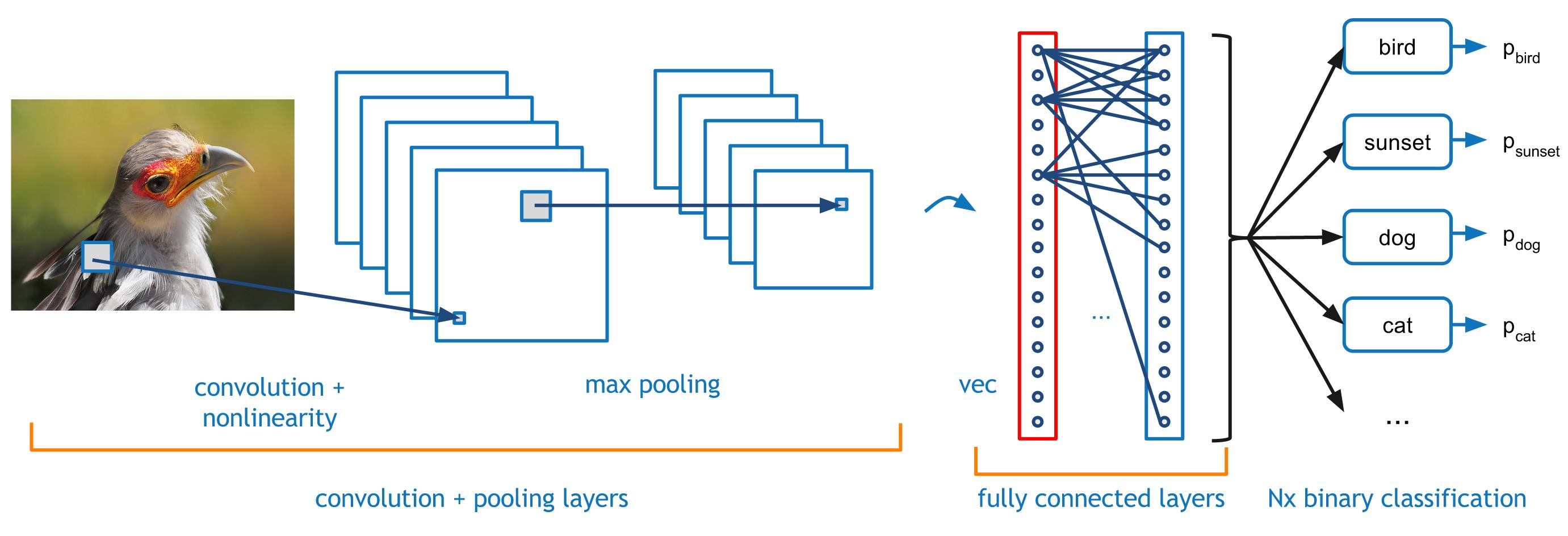}
    \end{center}
    \caption{Architecture of a CNN where the computation flow is explicitly shown -- Source: code.flickr.net}
    \label{fig:convnet}
\end{figure}

One particular architecture is the CNN, a special case of NN with certain restrictions.
CNN models learn very effective solutions for many Computer Vision problems in image classification.
The three key ideas of CNN are described below: local receptive fields, weight sharing and subsampling.
In CNN, a convolutional layer can model receptive fields by computing multiple trainable 2D kernels which is convoluted with its entire input whose result is called its feature maps.
A kernel convolution can be seen as the replication of a single unit along the dimensions of the input (a 2D grid for images), thus all weights are constrained to be equal by definition.
A convolutional layer contains multiple units where each has its own kernel.
Therefore different kernels are applied to the image where a single kernel convolution is called a feature map.
Such layer naturally computes filters which can be seen like data-driven feature extractors.
The benefit of weight sharing in convolutional layers is to reduce the number of global parameters to learn general purpose filters which can increase generalization.
Usually, CNNs have subsampling layers right after convolutional layers to reduce the dimensionality of the feature maps so that more concise and high-level information are extracted.
Most CNN architectures puts multiple convolutional layers connected to the input to extract visual features after which they have regular inner-product layers (see figure~\ref{fig:convnet}).
CNN can be seen as two parts: a trainable feature extractor made by the convolutional network followed by a neural network classifier.
Although CNNs have more constraints, it has been shown they generally outperform NN in multiple Computer Vision problems because they learn more invariance\cite{simard2003best}\cite{mnist_web}\cite{lawrence1997face}\cite{krizhevsky2012imagenet}.

Moreover, research found interesting structures in the last-layer embedding of classification NN using t-SNE.
The last layer tends to cluster samples of the same class together while separating the rest\cite{donahue2013decaf}\cite{yu2014visualizing}.
This is explained by the fact that deeper layers extract more high-level information, which is necessary to separate classes. Moreover the prediction is a direct product with the last layer, which encourages the network to have a simple structure directly clustered by classes.

\chapter{Theory: Dimensionality reduction}
\label{chap:dim_red}
Dimensionality reduction is an important method in machine learning that maps points in a high-dimensional space into a space with a lower number of dimensions.
This lower subspace is later called an {\em embedding}.
Representations suitable for human visualization should keep important relations between points of the original space.
In the experiments, we will apply dimensionality reduction on the neural networks' outputs.
This allows to quickly compare qualitatively the impact of controlled distortions applied on images over the networks' outputs.
The structure inferred by distances between points such as clusters, intra-cluster neighbors and outsiders, reveals important properties of neural network models.
Thus, accurate low-dimensional representations of embeddings that preserve local distances are desirable to compare different models.
In the following sections, dimensionality reduction methods are introduced where some are disregarded because they were not justified for this project, then the one used here, called t-SNE, is introduced in details.

\section{Optimization Problem}
Various algorithms differentiate themselves by several properties: their goal (\eg: interpolation, compression, visualization), by what they preserve (\eg: variance, distances), how they model point correlation (\eg: linearly or not) or whether they can model new points (\eg: a representation or a mapping).

First, the primary factor for the selection of a method is guided by the requirements, which is here to visualize.
Thus it should produce a 2D or 3D embedding, preferably in 2D, for easily interpretable scatter plots.
Secondly, we need an optimization problem that keeping certain relations between points such that clusters are equivalently represented.
This can be implemented by preserving relative distances to close neighborhood.
Considering the previous works discussed above, t-SNE is the best candidate in the current state-of-the-art.
The definition of the objective functions, the probabilities and the intuitive properties behind this method are now established.

Let us define the dataset $D$ formed by points in the input-space $X$ of dimension $N$: $D = \left\{ x_1, x_2, \dots, x_n \right\}$, each $x_i \in \R^N$, and one representation $D^\ast$ in the output-space $Y$ of dimension $M$: $D^\ast = \left\{ y_1, y_2, \dots, y_n \right\}$, each $y_i \in \R^M$.
The process of dimensionality reduction is to find the best representation $D^\ast$ that best preserves the most ``important'' information between $x_i$ and $y_i$ for each point.
SNE uses two Gaussian distributions for each point expressing the neighbor distances in $X$ and its equivalent in $Y$.
The Kullback-Leibler divergence is used to compute the objective function, which represents the mismatch of these two distribution for each pair:
\begin{eqnarray}
    L = \sum_{i=1}^n KL(P_i || Q_i) = \sum_{i=1}^n \sum_{j=1}^n p_{j|i} \log\left(\frac{p_{j|i}}{q_{j|i}}\right)
\end{eqnarray}
The probability for a pair of point $x_i$ electing $x_j$ in $X$ follows a Gaussian is as follows:
\begin{eqnarray}
    p_{j|i} = \frac{\exp(-|x_i - x_j|^2 / 2 \sigma_i^2)}{\sum_{k \not = i} \exp(-|x_i - x_k|^2 / 2 \sigma_i^2 )}
\end{eqnarray}
The equivalent probability $q_{j|i}$ in $Y$ is the same except that $\sigma = 0$ and $x$ is replaced by $y$.
Therefore, a closer pair implies a higher neighbor-election probability because the distance is low.
This cost function gives an asymmetrical importance to the distances: nearby points in $X$ are greatly penalized if they are far in $Y$; whereas a small cost is incurred for pairs far in $X$ but close in $Y$.

As said SNE suffers from crowdedness problems in the middle of the embedding and the optimisation is harder due to the asymmetrical nature of the objective function.
Both problems were addressed in t-SNE which give very good results in practice.
The two major differences with t-SNE is the symmetrization of the cost function and replaces the distributions of the embedding by student variants.
In t-SNE, a single objective function is minimized:
\begin{eqnarray}
    L = \sum_{i=1}^n KL(P || Q) = \sum_{i=1}^n \sum_{j=1}^n p_{ij} \log\left(\frac{p_{ij}}{q_{ij}}\right)
\end{eqnarray}
where:
\begin{eqnarray}
    p_{ij} = \frac{p_{j|i} + p_{i|j}}{2 n}
\end{eqnarray}
which forces outliers points to contribute more to the loss.
And:
\begin{eqnarray}
    q_{ij} = \frac{1 / (1 + |y_i - y_j|^2)}{\sum_{k \not = l} 1/(1 + |y_k - y_l|^2)}
\end{eqnarray}
which replaces the $q$ distribution in SNE by a Student with a heavier tail: distances in high dimensional spaces spread across more dimensions; in low dimensional space, the accurate equivalent distance needs to be much higher per dimension (thus more points end up further in $Y$, a heavier tail than in $X$).
As before, a point cannot elect itself: $p_{ii} = q_{ii} = 0$ and probabilities are symmetric for both distributions: $p_{ij} = p_{ji}$ and $q_{ij} = q_{ji}$.
As stated previously the input space $X$ has much more dimensions were distances can be expressed than the 2 dimensions of $Y$.
In summary, t-SNE uses Kullback-Leibler divergence to minimize the mismatch of these two spaces by means of probabilities, therefore the chance of important local structures (frequent patterns) being preserved is higher than with other methods (mosts do not express this goal through an objective function).
Global structures is encouraged by the coherence of the local structures as the divergence decreases and the system stabilizes.

\section{CNN for Dimensionality Reduction}
Usual dimensionality reductions like t-SNE are helpful for many visualizing tasks but it has important drawbacks as well.
The most important ones are: the computational cost, the incapacity of mapping new points and the indirect control over the resulting embedding.
Current implementations of t-SNE are still rare, unpolished and require tricks to make them tractable in practice (\eg: reducing first with PCA).
As there is currently no way with t-SNE to map new points (not in $D$), it is necessary to optimize the whole system from scratch.
Besides, t-SNE parameters like perplexity and learning rate are not simple to chose.
Fortunately there are new promising alternatives directly harnessing the power of NN.
Such models are introduced in the following section.

CNN are mostly used for classification but it can be optimized for other purposes as well.
Instead of the SoftMax loss function used in classification, a special training architecture with a suitable loss can be used to optimize topological constraints.
For example to create an embedding with particular properties in the output layer.
Moreover the ideas presented by reduction methods like t-SNE can be formulated in different terms to be applicable to NN.
In this work, the most important idea is to keep similar points together and dissimilar far away from each other.

\begin{figure}[t]
    \begin{center}
        \includegraphics[width=0.5\textwidth]{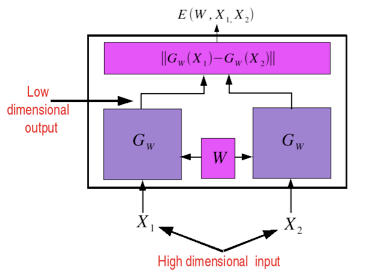}
    \end{center}
    \caption{Schematic representation of a Siamese network -- Source:~\cite{bromley1993signature}}
    \label{fig:siamese_network}
\end{figure}

The Siamese network combined with a contrastive loss is a good practical solution to train such networks\cite{bromley1993signature}\cite{chopra2005learning}.
Let us define $G_W$, the function that computes the network output, with parameters $W$ (see figure~\ref{fig:siamese_network}).
Then the Siamese network put two weight-sharing instances of $G_W$ side by side, each having their own input.
On top of this Siamese network is placed the ``cost module'', the contrastive loss, which will compute a loss proportional to the difference between the two output.
The complete network takes a pair of images as input, each image fed into a single $G_W$ instance, and the output is computed over their outputs.
The idea is to optimize the metric between points represented by $G_W$ and to spread contributions between the weights because they are shared.
At the end, when the network is used, only a single instance of $G_W$ is required to feed an input and $G_W$ gives the dimensionality-reduced point.

\begin{figure}[t]
    \begin{center}
        \includegraphics[width=0.5\textwidth]{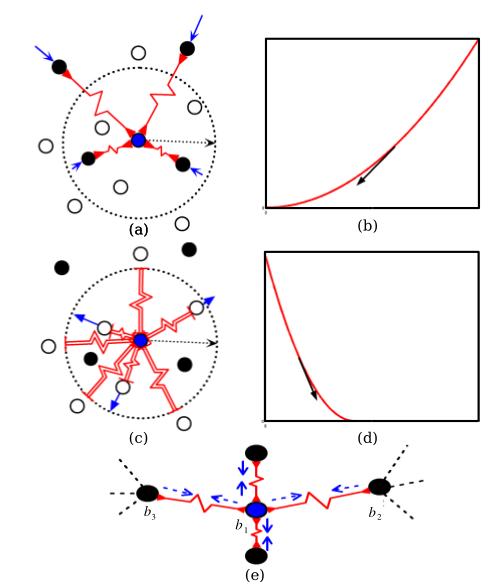}
    \end{center}
    \caption{A schematic representation of the contrastive loss using physical springs. Consult \cite{hadsell2006dimensionality} for more details. (a) shows the points connected to similar points with {\em attract-only} springs. (b) the loss function and its gardient associated with similar pairs. (c) the point connected only with dissimilar points inside the circle of radius $m$ with {\em m-repulse-only} springs. (d) shows the loss function and its gradient associated with dissimilar pairs. (e) shows the situation where a point is pulled by other points in different directions, creating equilibrium -- Source \cite{hadsell2006dimensionality}}
    \label{fig:contrastive_spring}
\end{figure}

The contrastive loss function takes its root from Energy-Based Models (figure~\ref{fig:contrastive_spring}).
The loss is constructed as follows: an attractive term is used to assemble similar points together.
Insufficient alone because it does not prevent degenerate solutions where all points are merged.
An opposing term is added to push dissimilar pairs to allow the system to converge towards a structured embedding.

More formally, the definition of the contrastive loss for a pair is as follows:
\begin{eqnarray}
    L = \frac{1}{2} Y (D_W)^2 + \frac{1}{2} (1-Y) \max(0, m - D_W)^2
\end{eqnarray}
where $Y \in \{0,1\}$ is the label: $1$ for similar pairs, $0$ otherwise; $D_W \in \R^N$ is the difference between the two networks' outputs and the parameter $m \in \R$ defines the minimal distance between dissimilar points.

In the contrastive loss, a label means whether a pair should be close in the embedding but the definition of similarity is left open for the application.

\section{CNN for Predictable Reduction}
The method above can learn an embedding which groups points or separates them based on the pairing strategy but there is still room for improvement.
The shape of the embedding is determined by the dataset labels which helps to form local structures.
However, there is no guarantee that this system will converge to an intended global structure with so many dimensions, such as: the ordering of the deformations (\eg: from lowest to highest) or their distribution (\eg: a line, plane or circle).
Secondly, the training required to make the embedding converge to a desirable structure can happen or not depending on the training time or specific initialization seeds, and this is partially due to the lack of constraints in the task.
When the number of embedding dimensions, $M$, is higher than the dimension of the pairing, $1$, the model can become unpredictable because more dimensions give more freedom in the way to represent these pairs.

To improve this situation, a solution is to add more constraints on such dimensions directly into the optimisation process.
We propose to give more than one information per pair and this allows to structure the embedding directly through the loss function.
This new method allows to control separately the usage of each dimension to express simultaneously different properties of the dataset.
To accomplish that, a generalization to the contrastive loss is proposed to work on training pairs with $p$-dimensional labels and an embedding with $M$ dimensions, where $p \leq N$ by definition.
In this framework, an embedding dimension is allocated to one particular component of the label and one component can be expressed through one or more embedding dimensions.
The allocation of the embedding dimensions and the pairing strategy are left open to the use-case.

This new loss function is now introduced with its formal definition.
Let us define $M$ as the number of embedding dimensions, $p$ the dimensions of the labels and $p \leq M$.
We define $D \in \N_+^p$ as the number of embedding dimensions assigned for each label component.
Only problems where each dimension is assigned to a single label and no dimension is left unconstrained are considered in this work: $\sum_i^p D_i = M$.
Then it follows that the definition of the generalized $p$-dimensional contrastive loss for a pair is:
\begin{eqnarray}
    L = \frac{1}{2} \sum_{i=1}^p \left( Y_i (D_{Wi})^2 + (1-Y_i) \max(0, m_i - D_{Wi})^2 \right)
\end{eqnarray}
where $Y \in \{0,1\}^p$ with $Y_i$ being the $i$th component of $Y$, $D_{Wi} \in \R^{D_i}$ is the difference between the two points in the sub-embedding for dimensions of the $i$th component, and $m_i \in \R$ is the minimal distance for dimension $i$.

\chapter{Methodology and Results}
\label{chap:results}

In this chapter, the following sections describe the general environment used to make the experiments, the settings to reproduce them and their outcomes.
The different models are introduced in details with their architecture, then the implementation using Caffe and Python is explained.
The generation of the datasets by combining samples from MNIST or NORB is also discussed.
Finally, the idea of quantitative measures is presented and used in the results for comparisons with the previous work.
The source code is available on GitHub at: \url{https://github.com/axel-angel/master-project}.

\section{Models and parameters}

As previously said, the work in the reference paper \cite{hadsell2006dimensionality} is closely followed here.
Therefore, we employ the same two models to experiment on the respective datasets MNIST and NORB.

\begin{figure}[t]
    \begin{center}
        \includegraphics{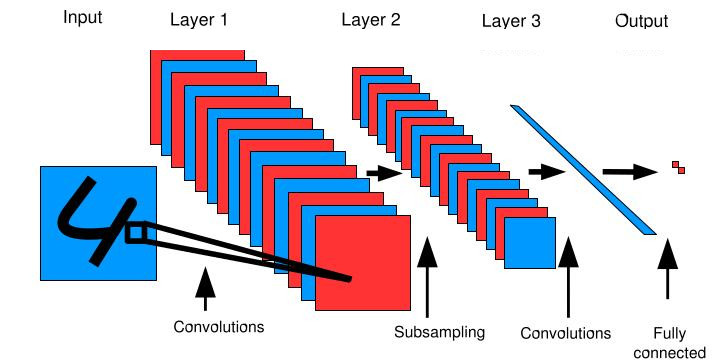}
    \end{center}
    \caption{Architecture of the CNN used for the experiments on MNIST (based on LeNet 5). The network contains two convolutions layer which are connected by a subsampling layer and the final output is computed by a fully-connected layer. The computation flow is illustrated for a region of an image. -- Source: \cite{hadsell2006dimensionality}}
    \label{fig:siamese_cnn}
\end{figure}

The {\bf first model} is {\em LeNet 5}, a multi-layer neural networks that is characterized by an architecture designed for handwritten characters as illustrated in figure~\ref{fig:siamese_cnn}.
In the experiments with the MNIST dataset, a variant is used with some minor changes described below.

This network architecture is comprised of a total of 7 layers with trainable parameters.
The first part of this network contains two pairs of convolution-pooling layers, where each convolution shares its weights defined as a kernel.
As explained earlier, each convolution is applied to the entire image to create one feature map and the ensemble of convolutions of a layer creates the feature map of this layer.
The first convolutional layer has 20 different trainable $5 \times 5$ kernels, which is followed by a trainable $2 \times 2$ max-pooling layer (with pixel stride of $2$).
The second convolutional layer has 50 trainable $5 \times 5$ kernels, followed again by a $2 \times 2$ max-pool layer (same stride).
These two pairs of layers form the convolutional network ({\em convnet}) part of this network which can be seen as {\em feature extractors}.
They will extract features encoding high-level cues (\eg: shape of strokes: straight or curved) in this application to discriminate between digits.

The second part of this network is made of two fully-connected layers which are connected through a non-linear activation layer.
The first layer has 500 trainable units computing an inner-product, followed by a non-trainable Rectified Linear Unit (ReLU) activations\cite{nair2010rectified}, followed by 10 trainable output units computing an inner-product which gives the digit class (1 versus rest).
These three layers form the neural network part used for classification based on the prior features.

The {\bf second model} is only made of two fully-connected inner-product layers.
This network is much simpler than the one above because it is trained on a subset of the NORB dataset, which exhibits very few variability compared to MNIST.
Indeed, only a single 3D object will be used and its shape in 3D is easily recognizable from any angle.
Back to the network, its two layers are made of 20 and 3 trainable units respectively, without non-linearity between them. 

In the following experiments, the variant of LeNet is trained in two different ways: for digit classification to analyze its ``natural'' embedding then with a Siamese training architecture with the contrastive loss function to analyze a ``constrained'' embedding.
In the second version, a single fully-connected layer is left in the NN part and reshape its number of units to match the embedding dimensions (\eg: $2$ or $3$).
The motivation is that the classification task is built on top of the network stack generating this embedding but in the second experiment this output needs to be projected, like in the reference paper.
Regarding this second model, a ``constrained'' embedding is directly trained with the Siamese architecture for NORB.

In the case of digit classification, the model is trained using Stochastic Gradient Descend (SGD) with a learning rate of 0.01 optimizing a SoftMax loss function.
Siamese models are trained with SGD with a learning rate of 0.01 or 0.001 optimizing the contrastive loss.

For this work, the {\bf Caffe} Deep Learning framework was chosen to make models.
There are several reasons such as: simplicity to express, train and manipulate networks and the inclusion of several practical architectures: LeNet on MNIST, Siamese on MNIST, AlexNet on ImageNet.
Moreover Caffe is well optimized in C++ for CPU and CUDA for GPU, flexible with official bindings for Python and MatLab and its community is very active and helpful.

The provided LeNet is easily adapted to match the architecture design discussed above for MNIST.
As the second model is very simple, it was trivial to write its definition ourself.
The training stage needs many important parameters related to the SGD solver: learning rate, batch size, momentum, weight decay (gamma and power) and number of epoch.
The default parameters were mostly used as defined in Caffe because the community already fine-tuned them.
The experiments are limited to $10'000$ iterations for MNIST and $100'000$ for NORB, with a batch size of $64$.
All networks were trained using the Caffe built-in solver started using the {\tt caffe train} command.
The alternative loss function was implemented in a new loss layer inspired by the built-in contrastive loss.
The implementation was straightforward to make in Python with NumPy and the Caffe bindings to integrate this layer into the network architecture.

The set of tools to generate distorted training sets were created ourself for the t-SNE experiments and to generate the distorted pairs training sets for the two Siamese experiments.
The scikit-learn Python library ({\em sklearn}) implements most of the standard algorithms for machine learning including: PCA and t-SNE\cite{pedregosa2011scikit}.
The performance of t-SNE depends heavily on the number of input dimensions and the usage of PCA on the datasets was necessary and followed other papers suggestions\cite{t-SNE}.
Likewise sklearn recommendation is to reduce to $50$ dimensions with PCA before applying t-SNE\footnote{t-SNE documentation page: \url{http://scikit-learn.org/stable/modules/generated/sklearn.manifold.TSNE.html}}.

For the visualization of the resulting embeddings, a solution based on the web library {\em CanvasJS} was developed for this project.
Its main advantage is to allow to interactively visualize with scatter plots directly from the output of t-SNE or the neural networks.
Several features were already provided: coloring points, fast plotting for interactivity but some were added by ourselves: zooming and moving the viewport, display the image of any sample and filtering/highlighting capabilities for the experiments.
All these functionalities provide the necessary tools to inspect the results and to come with the presented insights.

\subsection{Datasets}
To perform the experiments, two different datasets were used: {\bf MNIST}\cite{lecun1998mnist} (a popular handwritten digit dataset) and {\bf NORB}\cite{lecun2004learning} (a popular dataset of photographed 3D objects).
Each dataset is briefly discussed then a few characteristics is established regarding their variability and their usual usage in Computer Vision.

\begin{figure}[t]
    \begin{center}
        \includegraphics{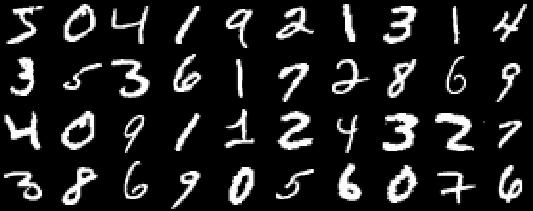}
    \end{center}
    \caption{A few samples taken from the MNIST training set to illustrate the dataset. Images were concatenated to save space.}
    \label{fig:mnist}
\end{figure}

The {\bf MNIST} dataset ({\em Modified National Institute of Standards and Technology}) is a gathering of multiple databases of handwritten digits.
One of the goal of MNIST is to provide a unified benchmark for digit recognition and it is widely used in Computer Vision for many years.
It is composed of $60'000$ training and $10'000$ testing images of digits between $0$ and $9$.
A few examples of the training set are illustrated in figure~\ref{fig:mnist}.
The samples are heavily post-processed: uniform black background, white-shaded digits with bold strokes.
The digits are centered such that the gravity center is in the middle and size-normalized so one digit lies inside a restrained sub-region of the $28 \times 28$ bitmap.
The variability of this dataset lies in the different strokes, handwritting style, natural rotations, thickness, curve roundedness and such.
The MNIST dataset is considered nowadays extremely simple due to the lack of natural variability and because state-of-the-art achieved extremely good results.
However it is still a good subject of experiments to try and validate new ideas.

\begin{figure}[t]
    \begin{center}
        \includegraphics{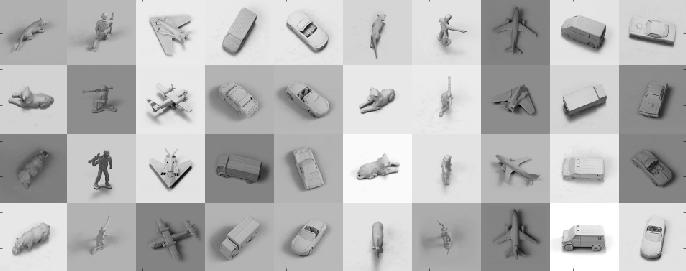}
    \end{center}
    \caption{A few samples taken from the NORB training set to illustrate the dataset. Images were concatenated to save save.}
    \label{fig:norb}
\end{figure}

The {\bf NORB} dataset ({\em NYU Object Recognition Benchmark}) is a collection of photos of toys taken from continuous poses.
This dataset has two variants and the normalized-uniform set was retained.
It is also intended for benchmark usages, in ``large-scale invariant object categorization''.
The set is composed of $48'600$ processed photos of toys, as illustrated by figure~\ref{fig:norb}.
The dataset is made by combining: $9$ elevation views, $18$ azimuth views, $6$ illumination conditions and $5$ toys categories, each containing $10$ toys.
The categories are: humans, animals, airplanes, trucks and cars.
The usual training task is to recognize the toy category whichever viewpoint or illumination is captured.
The DrLIM paper only selected a particular toy of a plane for its experiments and they infer the coherent representation of the camera viewpoint in 3D.
This same toy is used in this work to make the results comparable with DrLIM.

\begin{figure}[t]
    \begin{center}
        \includegraphics{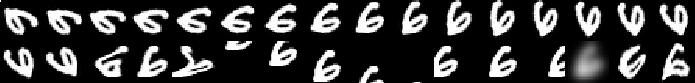}
    \end{center}
    \caption{Illustration of some distortions applied to a random sample included in the data-augmented dataset used in the t-SNE visualization experiment. The following distortions are shown: translations, rotations, blurring and linear deformation}
    \label{fig:mnist_transfo_tsne}
\end{figure}

\begin{figure}[t]
    \begin{center}
        \includegraphics{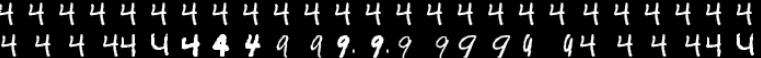}
    \end{center}
    \caption{Illustration of some pairs generated for a random sample with its distortions, neighbors and non-neighbors. The pairs are included in the training dataset for the Siamese network.}
    \label{fig:mnist_pairs}
\end{figure}

The experiments require to process the images in a systematic and coherent way.
Python scripts were used to automatically derive the training and testing sets in a predictable way based on the original dataset (MNIST and NORB).
The training architecture apply sequentially SGD over the samples in batch and loops over the whole dataset until the number of iterations reaches zero.
The generation of these datasets depends on multiple parameters related to the experiment.
In the case of the t-SNE visualization, a distorted dataset was created containing each sample plus its distorted versions varying in strength, where each image has only a single transformation at a time.
The samples are all distorted using the same set of intensities, quantified as follows: translations and shearing using pixel displacement, rotations using positive and negative angles and blurring with averaging radius (examples in figure~\ref{fig:mnist_transfo_tsne}.
In the case of Siamese training, paired datasets were created containing two distorted images with their similarity (1 or 0).
Images in a pair can have different distortions and different classes (digit for MNIST, elevation/azimuth for NORB) and still be paired (label 1), depending on the experiment.
An example of pairs for MNIST is illustrated in figure~\ref{fig:mnist_pairs}.

\section{Evaluation Metrics}
Qualitative results can provide insights regarding the model behavior or the quality of embeddings but it has drawbacks.
First it comes from human judgement which requires manual efforts, thus it cannot be systematic and automatic.
Secondly the subjective nature of qualitative measures is subject to different interpretation or can be biased towards certain aspect that people value differently.
We think qualitative judgements are unreliable and should not be used alone to asses the work in NN.
However, DrLIM results are lacking objective measures, for example the final loss or any sense of scale of dimensions to relate in the plots nor does it mention any kind of measure to quantify the quality.
By lacking any form of measures, we think that DrLIM paper does not promote the continuation of its work because there is no common way to measure ``good''.
Evaluation metrics are very important to permit objective comparisons between several results to see how they perform.

Thus we propose a very simple quantitative measure that allows to compare DrLIM and the new models.
Models are optimized with respect to their loss functions and indeed the loss is an objective measure that can be directly used to compare models on certain conditions.
Given the same loss, if its parameters are the same then it is logically fair to compare them.
Therefore, the contrastive loss of DrLIM is used on its test set using the original pairing strategy.
However, DrLIM paper does not provide the margin which is problematic to reproduce their work.
In the following, replicated versions of the DrLIM's models are trained ourself and thus a common margin is chosen per dataset for the two models.

\section{Results and Discussion}

This chapter is concluded by the presentation of the results and insights, including the plots and quantitative measures.
The first experiment is the visualization of the LeNet embedding on the MNIST classification task.
The structural differences are compared between models between trained with and without data-augmentation.
However the resulting embedding has an unexpected structures that does not fit the needed requirements.
Then, this method is left aside to learn directly optimized embedding with models using the contrastive loss.
The DrLIM's methods are applied by replicating the experiments on MNIST and NORB.
Followed by a throughout detailed discussion of the new methods in the same conditions to compare the two.


\subsection{t-SNE on LeNet}

In this experiment, two models for classification are trained on MNIST which is once unmodified, the second time it is data-augmented with translations.
The {\em data augmentation} process grows a dataset artificially by applying various distortions to train invariant models.
In this case, it contains each original digit plus many translated copies which are all labeled identically.
After the models are trained, their underlying embedding are inspected by using t-SNE.
The goal of this experiments is two folds.
First we would like to understand classifier embeddings and its interaction with t-SNE in a practical manner: testing various data-augmentation methods to see how it impacts the model, its accuracy and how t-SNE adapts.
Secondly, this experiment tests whether it is possible to get an embedding with the prerequisites properties of predictability with a state-of-the-art method specialized for projecting in 2D.

\begin{figure}[t]
    \centering
    \includegraphics[width=\textwidth]{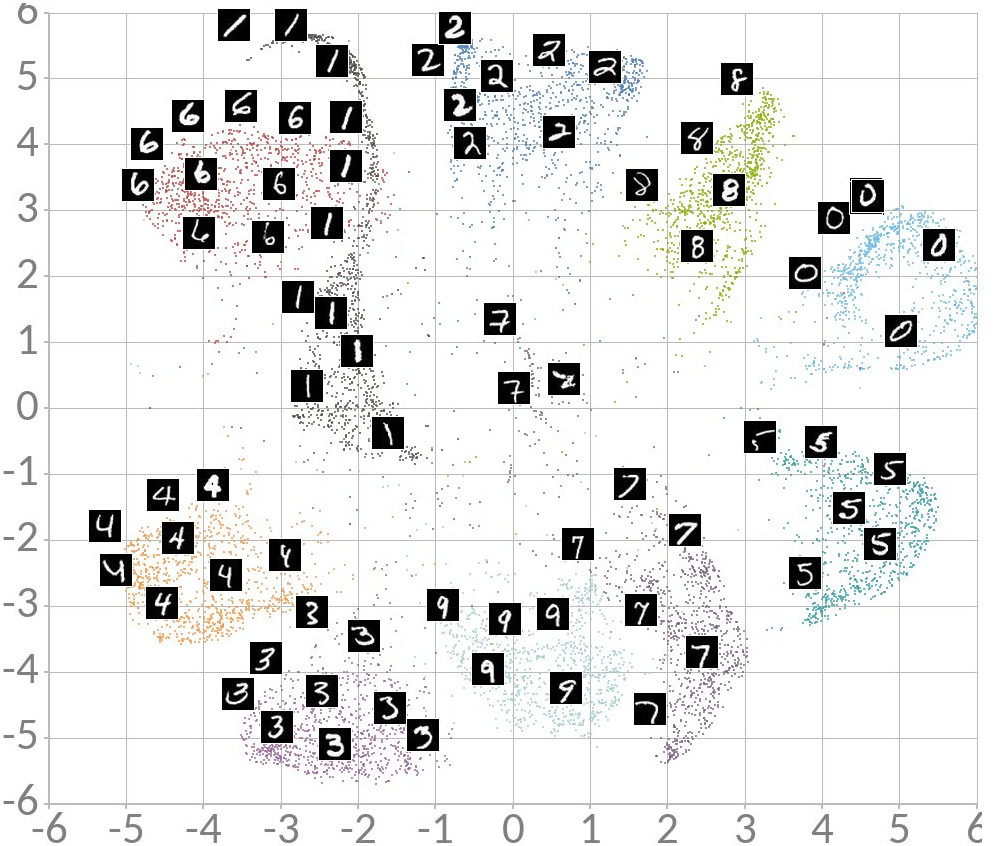}
    \caption{The t-SNE representation of the CNN embedding for the MNIST test set using the interactive tool.
    This figure shows a general view of the non-distorted samples.
    Some random samples are shown for each cluster.
    Each color represent a digit class.
    The original digits are well separated into clusters for each class.
    The major exception is the 1s spreading on a thin vertical curve over a small part of he 6s.
    Inside clusters, digits are spread coherently to their characteristics.
    }
    \label{fig:mnist_nda_tsne}
\end{figure}

\begin{figure}[t]
    \centering
    \includegraphics[width=\textwidth]{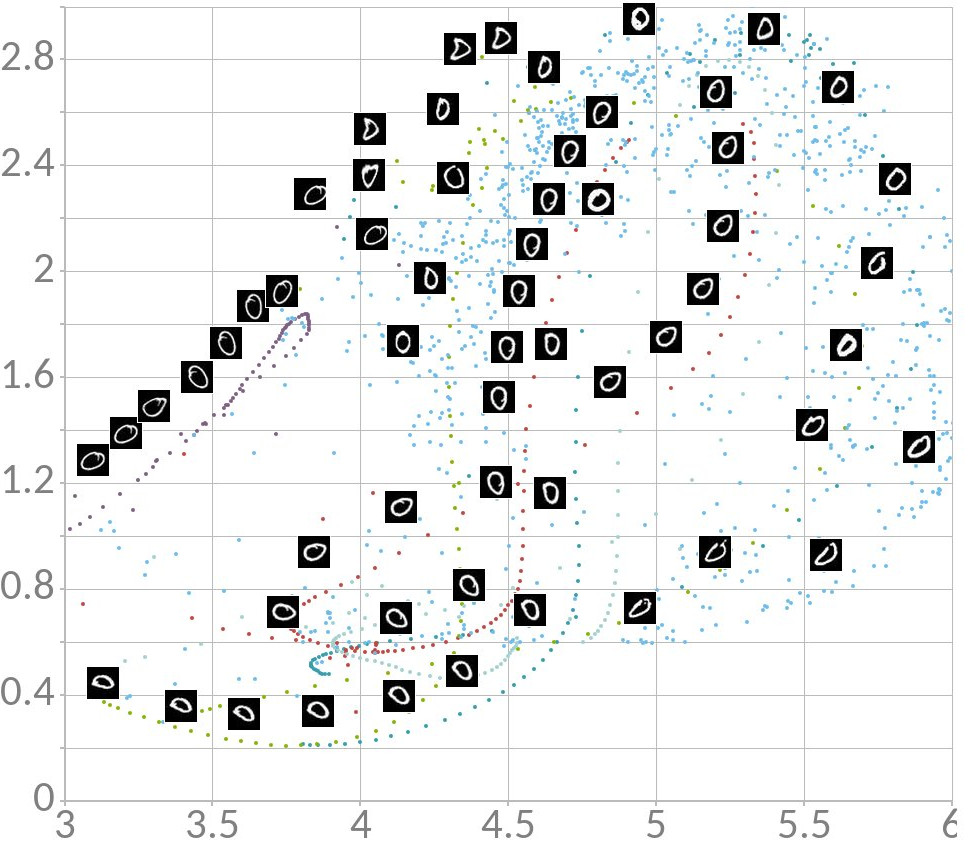}
    \caption{The t-SNE representation of the CNN embedding for the MNIST test set using the interactive tool.
    This figure shows a close view of the cluster for 0s including this time the distorted samples.
    A unique color is given to each digit class as before plus one unique color for each sample's distortions.
    Translated samples quickly end up outside the digit cluster whereas other distortions like rotations are structured into curved lines.
    Most illustrated images are rotations of a few selected samples.
    }
    \label{fig:mnist_nda_tsne2}
\end{figure}

To create the first model, the training is done on the MNIST classification task to predict digits without any modification.
At this point, all images of the training set are used without data-augmentation, thus there is little invariance.
The error rate of this model is 1.0\% which is computed over the MNIST test set.
According to the published results on the MNIST website, this is a reasonable value (LeNet-5: 0.95)\cite{mnist_web}.
In the following visualization with t-SNE, the classification layers of this this network are removed (SoftMax and 10-outputs inner-product).
Therefore the last remaining layer is a ReLU which follows the last inner-product of 500 dimensions.
The hypothesis at this stage is as follows: this last layer expresses high-level concepts of digits.
To understand the behavior of the system, a few distorted samples are included in the test set: for each digit, five samples are selected and their distortions included.
This modified test set is feed-forwarded into the network, apply PCA, then apply t-SNE to create the final embedding.
The whole process takes around 45 minutes on a single core.
The resulting embedding is illustrated in the figures~\ref{fig:mnist_nda_tsne} and ~\ref{fig:mnist_nda_tsne2}.
Most of the original digits are well separated on the first figure where each cluster represents a single class.
The major exception is the 1s spreading on a thin vertical curve over a small part of he 6s.
The hypothesis seems validated by the structure on this plot: class separation and coherence inside clusters.
Most of the distorted samples on figure~\ref{fig:mnist_nda_tsne2} are part of their respective clusters however they start to drift towards other classes as their distortion increases and the most extreme examples are at the opposite side.
At this point, it should be noted that the model is very sensitive to translations and many reside in the wrong clusters.

\begin{figure}[t]
    \begin{center}
        \includegraphics[width=\textwidth]{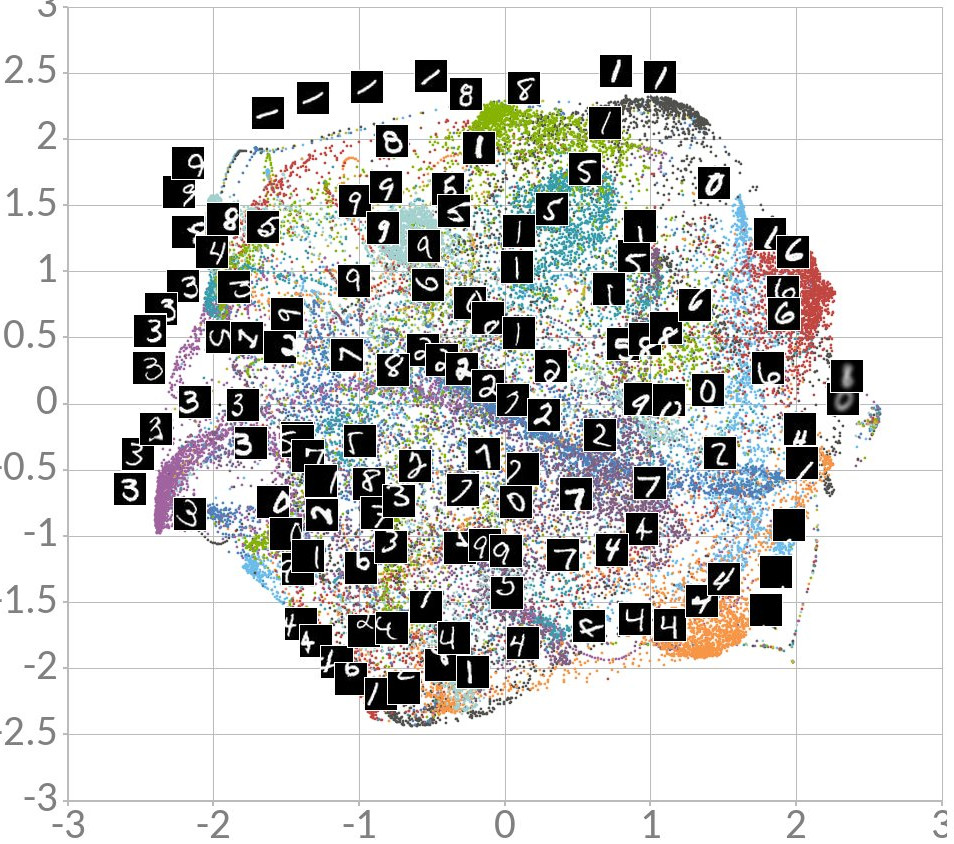}
    \end{center}
    \caption{The t-SNE representation of the CNN embedding for the data-augmented MNIST test set using the interactive tool.
    A unique color is attributed to each digit class and one for each sample's distortions (as before).
    The illustrated digits were manually picked to represent local trends.
    Global structures are dominated by translations but most of the translated digits are grouped disregarding their class.
    Local structures exhibit partial coherence by packing digit classes together but the overlapping is too important to ignore.
    }
    \label{fig:mnist_da_tsne}
\end{figure}

In the next experiment, the same process is applied with a translation-invariant model.
One simple way to achieve invariance is to train the model over a data-augmented training set to classify correctly translated digits as well.
The augmentation consists of a range of translations applied to all samples which are included into the training set.
In this case, all translations in $x$ between $-5$ to $+5$ plus translations in $y$ between $-10$ and $+10$ are generated.
The error rate on the original test set increased up to $3.5\%$, but there is no comparison for this value.
However this model has a strong translation invariance because its error rate is only $8.5\%$ on the data-augmented test set (it contains extreme translations).
The second hypothesis is as follows: As the features contain many different characteristics, it should also detect translations which was heavily present in the training set.
The embeddings of this new model can be now compared (see figure~\ref{fig:mnist_da_tsne}) versus the one above.
From a general perspective, translations are dominating the global structure of the clusters.
Many centered classes are still in their own clusters but most of the translated digits are grouped disregarding their class.
However the ``translated'' clusters exhibits coherence: sub-clusters represent digit classes but they are heavily overlapping.
For example, the right-displaced digits are exclusively appearing in the top-left corner where all digits are present and mostly stacked together in the center.
This is actually a general problem with this plot: many clusters appear but their delimitations are fuzzy which suggests it would be hard to quantify translations and digits.

From these two experiments, important knowledge emerges about this classification tasks, the data-augmentations and t-SNE as well.
Manual experiments confirmed that LeNet is sensitive to the employed distortions and can easily misclassify if no form of data-augmentation is used during training.
As previously said, nearly no natural distortion is present in the original dataset and this is partly due to the heavy prior processing applied to the MNIST dataset.
However small rotations are still present in MNIST which explains why the new model is less prone to errors on artificially rotated samples.
Data-augmentation is an effective way to reduce the error rate on certain type of distortions but it has a cost.
In this case, t-SNE embedding looks ``overcrowded'' and most clusters are not well separated anymore.
We think it is due to the important difference of energy in the pixel space, \eg: the center of mass is translated.
One explanation is that the network learned to separate this source of variance and acquired position-specific features to distinguish translated digits which impacts characteristics of the last layer.
Then t-SNE is impacted by clustering translations first, then digits as its second major structure.
One important insight is that t-SNE seem to represent hierarchical structures with decreasing importance thus the most discriminative features are weighted more.

These results are unexpected: the clusters are not well separated and they are lacking in coherence even though the task require to separate the digits.
This experiment was meant to create an invariant model for classification but the representation does not naturally split digits and translations into different axes.
It should be noted again that t-SNE is oriented towards visualization and the goals of visualization are not formally and universally well defined.
The current formulation of this method is to preserve distances between the features which is not sufficient for the application.
There is currently no way to add prior information like supervised methods to guide the optimization to associate particular dimensions to particular variance in the data.
A possible solution is to modify the objective function of t-SNE but it has the important limitations previously discussed.
The lack of end-to-end optimization in this method is also important and this can be solved by using layers handling the dimensionality reducing into a single NN.

In the following part, results of an alternative method is presented based on end-to-end and supervised training of a single CNN.

\subsection{Contrastive LeNet (MNIST)}
In the two following experiments, Siamese models were trained based on DrLIM settings on both MNIST and NORB so that the results are comparable between the reference paper and the networks with the extension as described below for two-label pairs.
The goal of DrLIM is to train a model where points are close together if they are ``visually'' similar without considering distortions depending on the dataset.
The implementation of their pairing strategy is as follows.
First they group each sample into a common ``neighborhood'' relation: the 5 most similar ones using the Euclidean distance in pixel space.
However in practice, neighbors with different digit class would appear and DrLIM's paper did not mention if they considered legit or not.
In this work, only neighborhood with the same class is used because a manual look unveils that they are in fact quite dissimilar.
The sample and its $n$ translations are all paired with both its 5 neighbors and all its $n$ translations (this sums up to $n \times 5n$ pairs per sample).
All other pairs are considered dissimilar: when the two samples are not part of the same common translated neighborhood.
This strategy uses the regular contrastive loss function in 1D with a single similarity per pair.

The objective is comparable in the sense that staying invariant is desirable but only in certain dimensions priorly picked.
The new implementation can still quantify distortions by expression them in the other dimensions.
One important advantage is to allow more control, especially in how dimensions are used.
As said, a single model is trained which is less expensive and it learns reusable features for multiple pairings.
The extension of the contrastive loss function can be used with $p$ labels for each pair.
The definitions of $p$ and $D$ for the two Siamese experiments are described and justified later.

In the case of MNIST, a 3-dimensional embedding space is used: where the first two dimensions express neighborhood similarity and the last dimension expresses the distortion similarity.
This case only needs two dimensions but to make it comparable to DrLIM's 2D embedding, a 2D neighborhood space will be used as well and distortions are separated into its own space.
Making comparisons between 2D space (DrLIM) and 1D neighborhood (this work) would be unfair due to distances increasing very quickly as the number of dimensions increases.
In this settings: $M=3$ (embedding dimensions), $p = 2$ (two labels) and $D = \left\{ 2, 1 \right\}$ (2D + 1D).
The network is trained using a pair strategy inspired by DrLIM but with important differences.
The first label is called the ``neighborhood'' similarity and considers: a sample with its translations to be similar, the sample with its neighbors also similar if they have the same translations, and the sample with any non-neighbor sample to be dissimilar with or without translation.
The second label is called the ``transformation'' similarity and considers: a sample with any other sample with the same translation similar, and any pair without the same translation dissimilar.
Pairs not mentioned in this descriptions above are not included in the datasets.
Moreover the size and the time for training is reduced with a low impact by randomly taking a subset of dissimilar pairs to balance the label ratio instead of including all of them.

In the case of NORB, a 3D embedding is used as well.
The first two dimensions are allocated for the cyclic azimuth (horizontal angle of the viewpoint) and the last dimension for the non-cyclic elevation (vertical angle).
DrLIM's work defined similar pairs when the two images are from contiguous elevation or azimuth, this work will continue likewise.
The justification of DrLIM's dimension allocation is as follows: the azimuth viewpoint is cyclic and the shape, to represent such a structure without overlapping, is an ellipsis and this requires at least two dimensions.
The elevation viewpoint is not cyclic because NORB has only a smaller subset of values, thus only a single dimension is allocated.
A direct comparison with DrLIM's embedding is possible in this case as the models work on the same problem.
Once again: $M=3$ (embedding dimensions), $p = 2$ (two labels) and $D = \left\{ 2, 1 \right\}$ (2D + 1D).
The network is trained based on DrLIM's pairing strategy where the azimuth and elevation are separated into two labels (instead of having them merged).

\begin{figure*}[h]
    \centering
    \begin{subfigure}{0.45\textwidth}
        \centering
        \includegraphics[width=\textwidth]{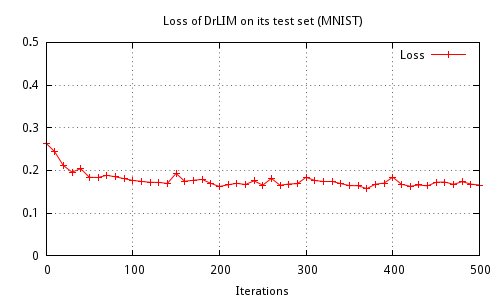}
    \end{subfigure}
    \begin{subfigure}{0.45\textwidth}
        \centering
        \includegraphics[width=\textwidth]{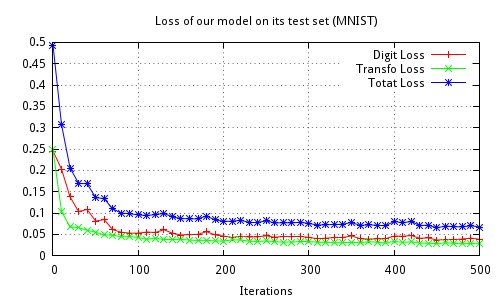}
    \end{subfigure}
    \caption{The model losses for 500 iterations of DrLIM and the new model on MNIST.
    The two losses cannot be compared directly but the speed of convergence and the final difference are key.
    Both models show sign of convergence after $100$ iterations but the new model has reached a simpler solution due to the different pairing strategy chosen in this work.
    An important point to emphasis is that DrLIM is in 2D whereas the new model is in 3D.
    }
    \label{fig:mnist_cl2d_loss}
\end{figure*}

\begin{figure}[h]
    \centering
    \includegraphics{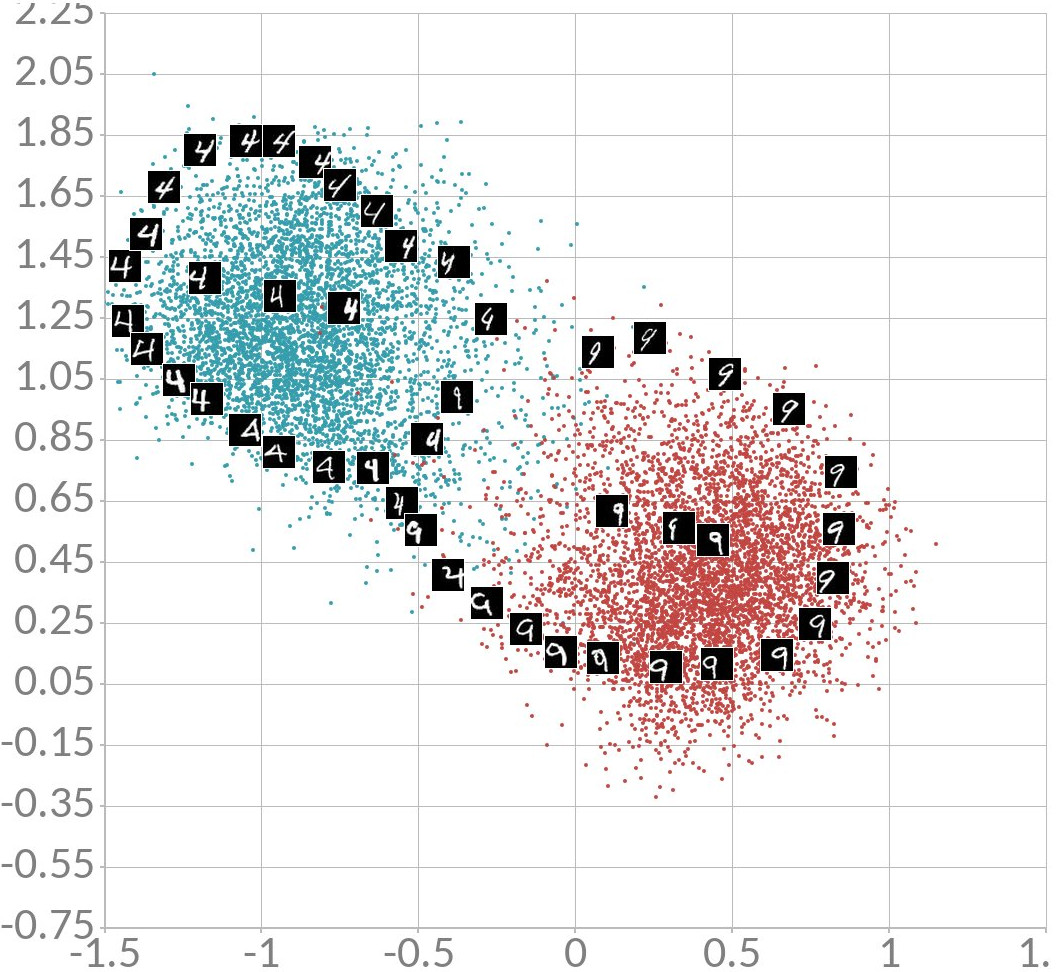}
    \caption{DrLIM embedding for the MNIST translation-augmented test set where some random samples are shown for illustration.
    This figure is very similar to the own present in DrLIM paper and also exhibits coherent groupings based on translation-agnostic similarity.
    }
    \label{fig:mnist_cl_drlim}
\end{figure}

\begin{figure}
    \centering
    \includegraphics[width=\textwidth]{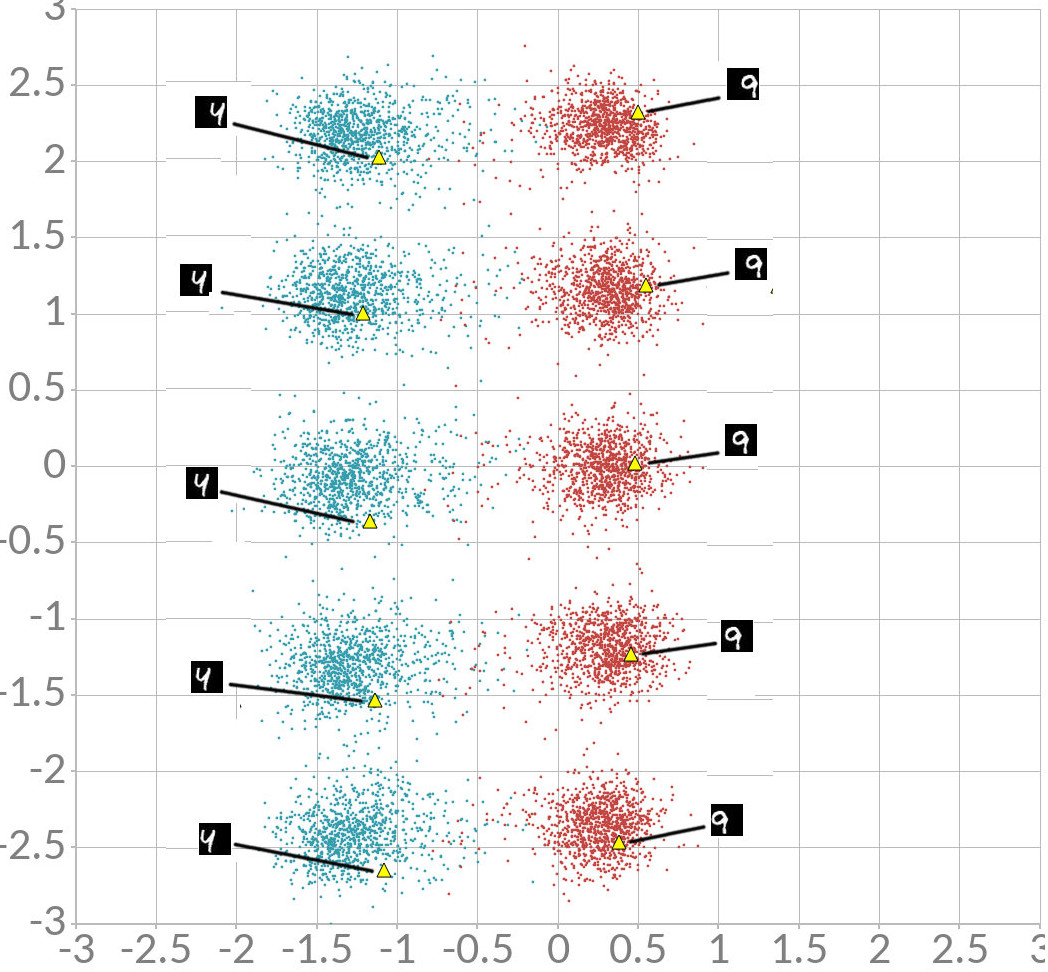}
    \caption{The new model embedding on the MNIST translation-augmented test set projected into its last two dimensions.
    Two samples of different class were picked at random and their distortions are marked as triangles in the figure.
    This 3D embedding has an additional vertical dimension compared to DrLIM that quantifies the displacement and the other top-down projection of this embedding would give an identical plot as above for DrLIM.
    Clusters are formed for each translation intensity and each digit class in a well separated manner and remarkably they are logically ordered.
    }
    \label{fig:mnist_cl2d_1}
\end{figure}

\begin{figure}
    \centering
    \includegraphics[width=\textwidth]{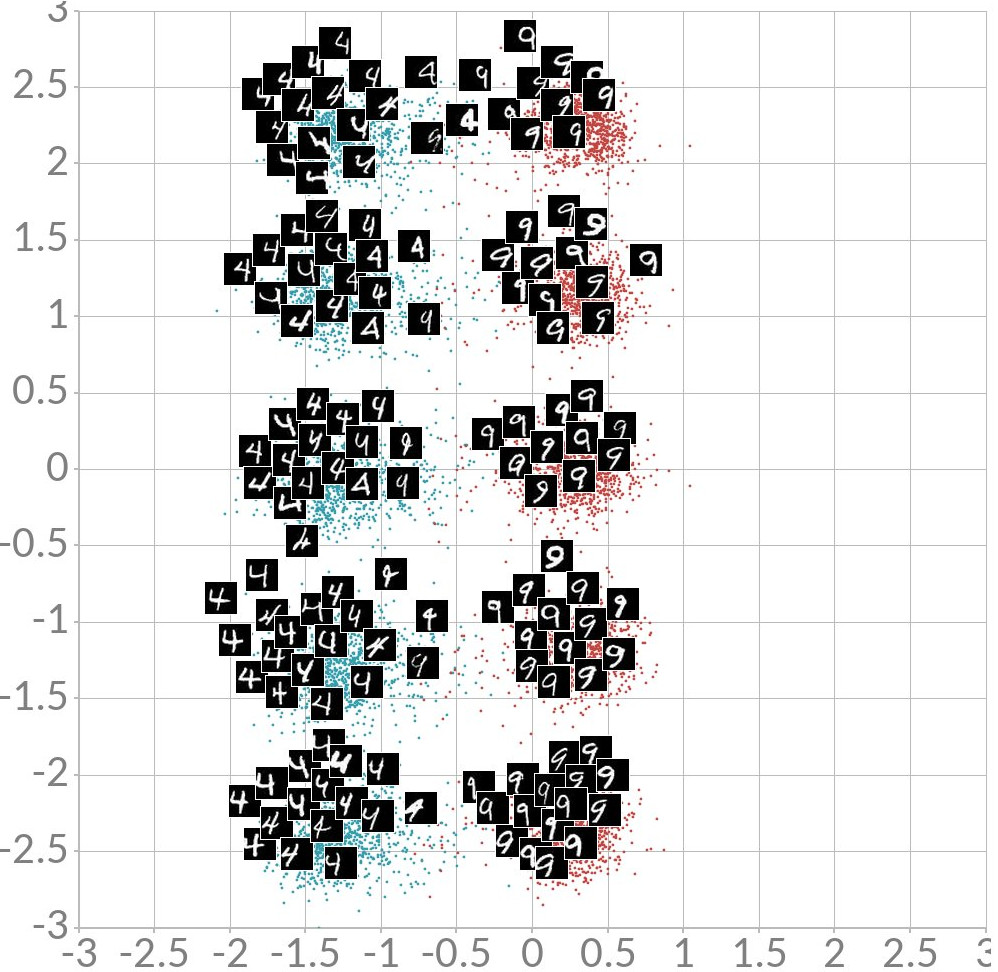}
    \caption{The new model embedding on the MNIST translation-augmented test set projected into its last two dimensions.
    General view where many random sample's distortions are displayed to represent the local tendencies.
    The space inside each cluster appears to be also coherent.
    }
    \label{fig:mnist_cl2d_2}
\end{figure}

The first dataset is a subset of MNIST containing only 4s and 9s digits which are used to derive the training and testing set based on the 2D pairing strategy.
Therefore the resulting dataset is composed of each sample plus its translations (±3, ±6 pixels) which is then paired.
The models are trained until the loss on the respective test set has attained a local minima, which takes around 10 minutes.
After around $100$ iterations, both models have reasonably converged to a stable solution (see figure~\ref{fig:mnist_cl2d_loss}).
However, the new model has reached a lower solution than DrLIM: comparing them directly is not possible because the two datasets are different but the difference in the loss in this model is more important and converge as quickly.
Intuitively this means that the new problem formulation is easier to solve and the model finds a ``better'' solution, in the sense it could satisfy more the new constraints described by the loss.
The resulting embeddings are shown in figure ~\ref{fig:mnist_cl_drlim} for DrLIM and figures ~\ref{fig:mnist_cl2d_1} and ~\ref{fig:mnist_cl2d_2} for this model.
The DrLIM embedding reproduced in this work is very similar to the expected result: a coherent grouping based on translation-invariant similarity but the separation between digits is more pronounced.
The DrLIM embedding can be considered as a top-down projection of the new 3D embedding where the additional vertical dimension quantify the displacement.
The different intensity of translations are well separated into their own cluster as demonstrated by a meticulous manual inspection and remarkably they are ordered vertically by their intensity.
Once again, the space inside clusters is well organized by digit features like DrLIM.

\begin{figure}[h]
    \begin{center}
        \includegraphics{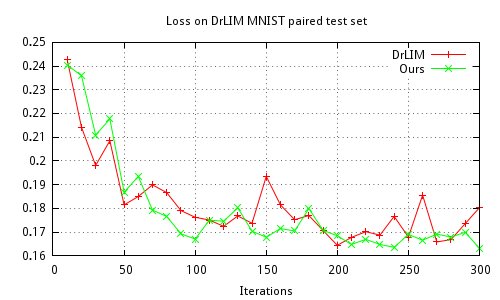}
    \end{center}
    \caption{Comparison of quantitative measures between DrLIM and the new model.
    The measures are taken during training iterations over the translation-augmented MNIST test set.
    The two loss series are statistically indistinguishable and it can be concluded that both models learned an equivalent representation whereas the new model can also expresses distortions without noticeable impact.
    }
    \label{fig:loss_mnist_test_common}
\end{figure}

In order to compare the two methods, quantitative measures are presented where the contrastive loss is computed on the DrLIM paired test set as described in the methodology.
The evolution of this measure during training is plotted in figure~\ref{fig:loss_mnist_test_common} and the final losses after 10'000 iterations are: $0.160023$ (DrLIM) and 0.145418 (the new model).
The two loss series shown in this figure are statistically indistinguishable according to a t-test with a null hypothesis of $\alpha = 0.05$.
Therefore it can be concluded that the new model learns an equivalent representation in its 2D digit-projection to DrLIM with a slightly different formulation and it can expresses distortions without loosing information in its 3rd dimension at the same time.

\begin{figure}[h]
    \begin{center}
        \includegraphics{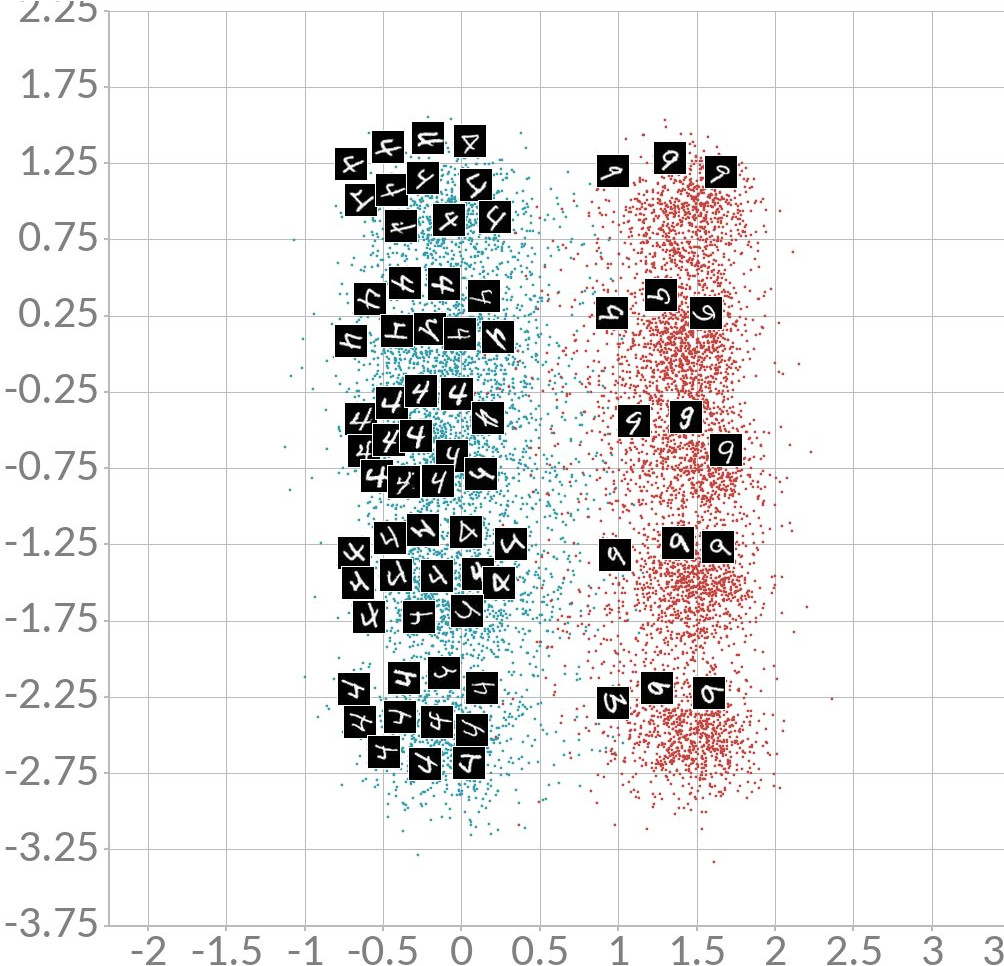}
    \end{center}
    \caption{The new model embedding on the MNIST rotation-augmented test set.
    Clusters are remarkably arranged in a linear fashion although rotations are non-linear.
    }
    \label{fig:mnist_cl2d_rotate}
\end{figure}

The new model was tested on a data-augmentation of MNIST based on rotations as well to show it works on non-linear deformations as well.
Clusters are still arranged in a linear fashion although rotations are non-linear but the cluster separations are smaller, see figure~\ref{fig:mnist_cl2d_rotate}.
The results are similar to the previous experiment and the figure is provided for demonstration as it will not be discussed further.

\subsection{Contrastive LeNet (NORB)}
\begin{figure*}[h]
    \centering
    \begin{subfigure}{0.45\textwidth}
        \centering
        \includegraphics[width=\textwidth]{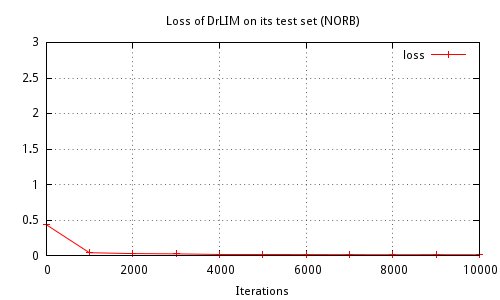}
    \end{subfigure}
    \begin{subfigure}{0.45\textwidth}
        \centering
        \includegraphics[width=\textwidth]{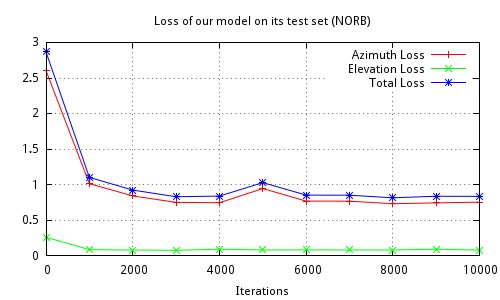}
    \end{subfigure}
    \caption{The model losses for 10'000 iterations of DrLIM and the new model $m=10$ on NORB.
    After $2'000$ iterations, both models have reasonably converged to a stable solution.
    DrLIM model has a very low loss and the new model faces problems to correctly represent the azimuth dimensions.
    }
    \label{fig:norb_cl2d_loss}
\end{figure*}

\begin{figure*}[h]
    \centering
    \begin{subfigure}{0.65\textwidth}
        \centering
        \includegraphics[width=\textwidth]{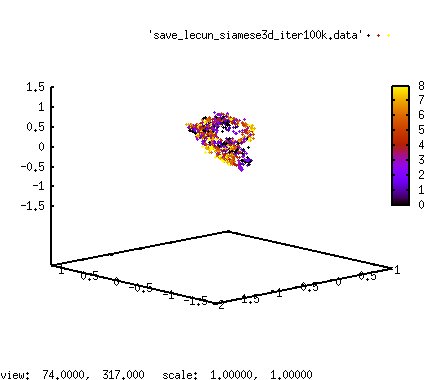}
    \end{subfigure}
    \begin{subfigure}{0.65\textwidth}
        \centering
        \includegraphics[width=\textwidth]{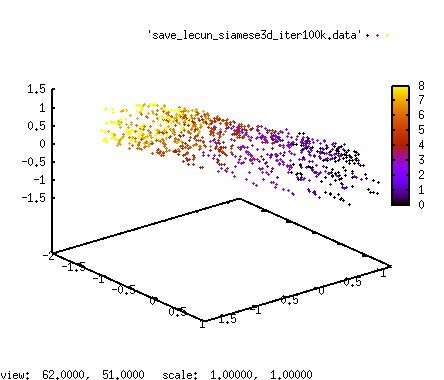}
    \end{subfigure}
    \caption{The DrLIM embedding in 3D on the NORB 1-plane complete dataset after $100'000$ iterations.
    The colors indicate the elevation angle.
    The result found that a thin cylinder whose long axis represent the elevation and the radius is the azimuth angle.
    Such results are remarkable because the model found this shape on its own.
    However it is not perfect, some parts are flat along the elevation axis and the radius is fuzzy.
    }
    \label{fig:norb_drlim_embedding}
\end{figure*}

\begin{figure}
    \centering
    \includegraphics[width=\textwidth]{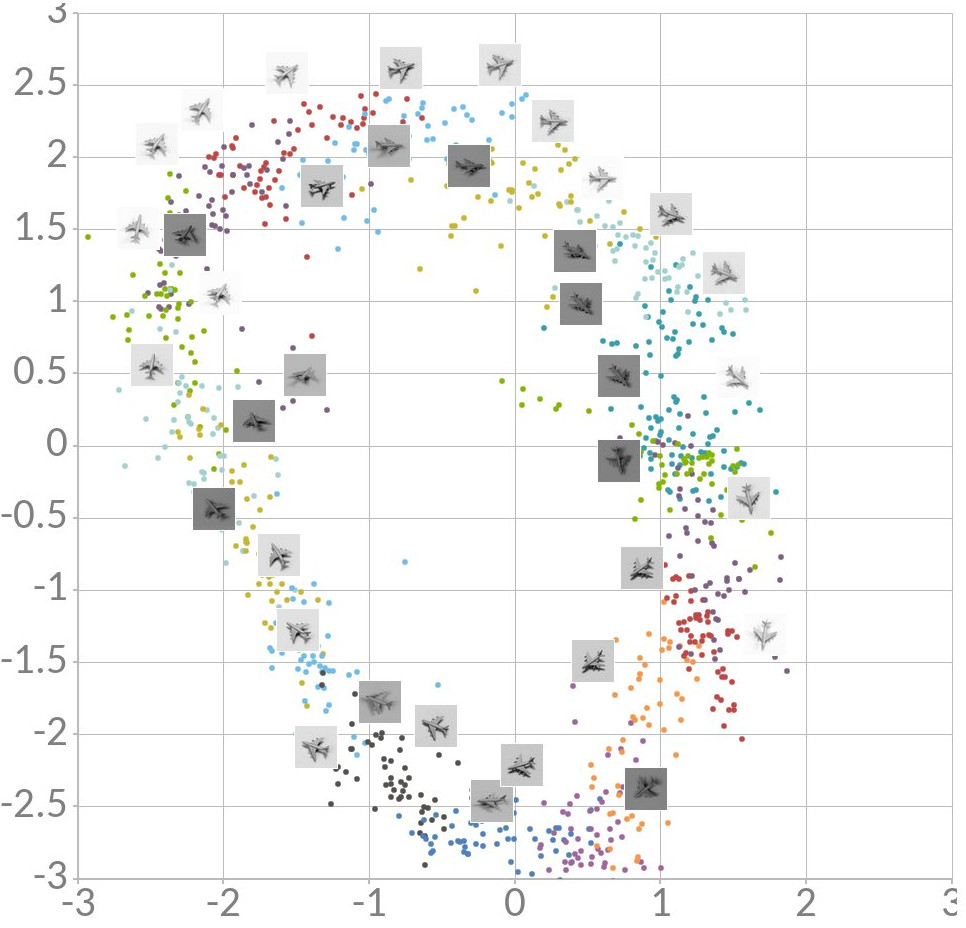}
    \caption{The new model embedding on the NORB 1-plane complete after $100'000$ iterations.
    2D projection of the first two axes.
    Colors indicate the azimuth angle.
    The major difference with DrLIM is the alignment along the plot axes and a better separation inside the cylinder.
    The mapping is also invariant to the lighting conditions.
    }
    \label{fig:norb_cl2d_embedding_1}
\end{figure}

\begin{figure}
    \centering
    \includegraphics[width=\textwidth]{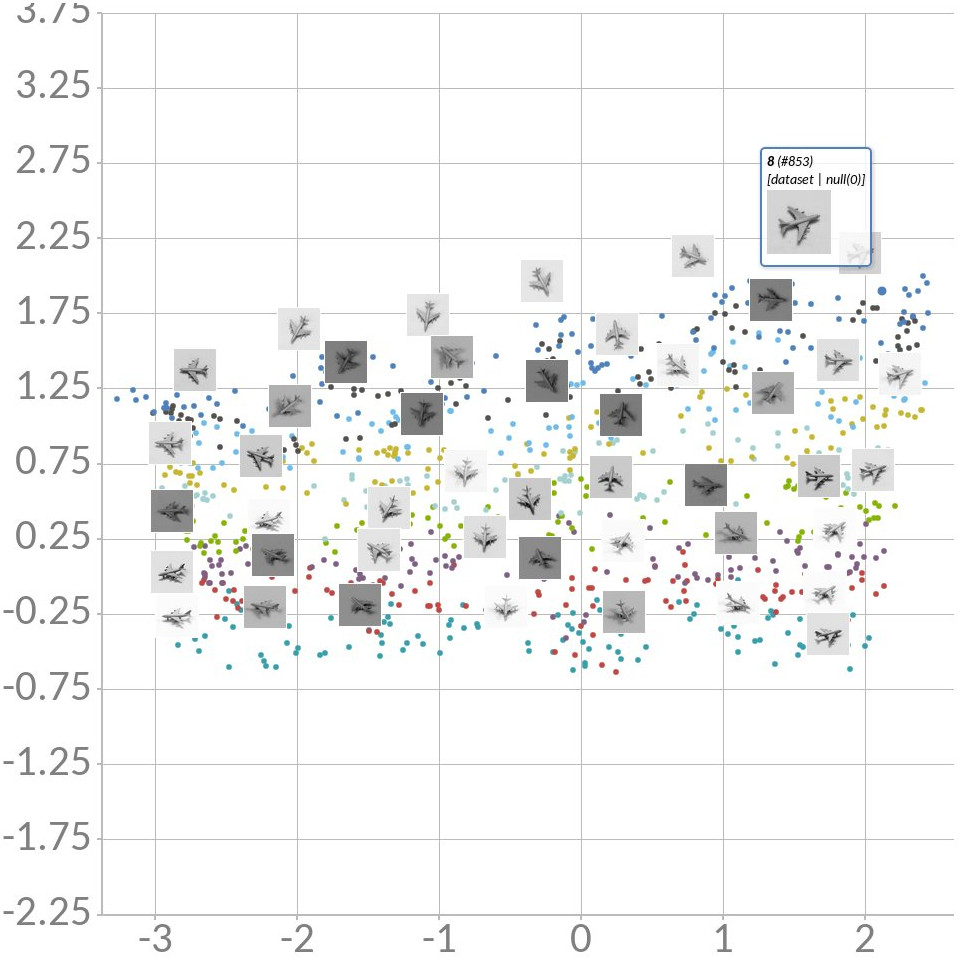}
    \caption{The new model embedding on the NORB 1-plane complete after $100'000$ iterations.
    2D projection of the last two axes.
    Colors indicate the elevation angle.
    The relation between elevation and azimuth is well organized in the embedding.
    }
    \label{fig:norb_cl2d_embedding_2}
\end{figure}

Once again, the network is trained following the new conventions, this time on a subset of the NORB dataset with a single plane.
As previously described, this subset is composed of all viewpoint of a single plane which is later called the 1-plane NORB dataset.
This subset is split into two parts: 660 training samples and 312 test samples.
These two sets are both separately paired according to the strategy described in the methodology to form the training and testing sets.
The models are trained until the loss on the respective test set has attained a reasonable local minima, which takes around 10 minutes.
After $2'000$ iterations, both models have reasonably converged to a stable solution (see figure~\ref{fig:norb_cl2d_loss}).
To get the most convincing results, a margin $m=1$ was picked for DrLIM and $m=10$ for the new model, therefore their loss on the test set should not be compared directly, instead the quantitative measures below should be used instead.
However, it can seen that DrLIM in figure \ref{fig:norb_drlim_embedding} has an easier formulation this time (smaller margin) and the embedding is very similar to the one presented in the reference paper.
The most important features of the embedding to represent NORB are: the cyclic structures for the azimuth angle, the continuity along the two axes disregarding lighting illumination and the sharp separation along the axes.
Indeed, this embedding is a thin cylinder whose long axis represent the elevation and the radius is the azimuth angle.
However, the cylinder is not perfect: multiples parts are more flat along the elevation axis (like a plane) and the radius is fuzzier than presented in the reference paper.
The major difference with the new solution, as shown in figures \ref{fig:norb_cl2d_embedding_1} and \ref{fig:norb_cl2d_embedding_2}, is the alignment along the axes, a more stable radius and a better separation inside the cylinder.
The predictability of the solution is improved by these properties compared to DrLIM.

\begin{figure}[h]
    \begin{center}
        \includegraphics{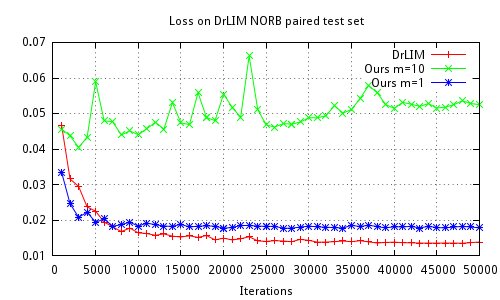}
    \end{center}
    \caption{Quantitative measures to compare between DrLIM and the new model during training iterations on the 1-plane NORB test set.
    The $m=10$ model put less importance to the proximity of similar neighbors than the other two models.
    This plot suggests that the $m=1$ model is better but $m=10$ is qualitatively better, although it has a higher loss.
    The loss function does not seem to reflect the visual quality in this case.
    }
    \label{fig:loss_norb_test_common}
\end{figure}

To compare the two models, the contrastive loss was computed on the test set of the 1-plane NORB with the same parameters ($m=1$) as with MNIST.
The loss evolution is shown in figure \ref{fig:loss_norb_test_common} where a version of the new model was also added with $m=1$.
The embedding of this model has many aspects of the $m=10$ model and the main difference is instead of having a round shape for the azimuth, it is similar to a heart which is also cyclic but less desirable.
A major difference can be found with the $m=10$  model compared to both DrLIM and the $m=1$ model.
By definition of the loss function, the model is not penalized to put dissimilar pairs further than the margin, therefore it means the $m=10$ model has decided it is better to put similar neighbors further than the other two models.
Moreover, the loss suggests that the $m=1$ model is superior but $m=10$ looks better, although it has a higher loss.
Unfortunately, the visual quality is not reflected by the loss function in this case.

\subsection{Remarks}
To conclude this chapter, a few remarks about the general results and some difficulties during the experiments should be noted.
The primary goal of the experiments using contrastive losses is to show more predictability can be expected when given more control, over each dimension and over the clustering, with a simple extension of the original formulation.
Our qualitative judgement is positive about the success of the experiments.
The first experiment demonstrated that models based on DrLIM can include more informations concerning the distortions with a minor impact over the original feature space.
In our second experiment it was shown that these models can also be guided to represent the NORB space by using dimensions in a certain way with minimal efforts.
Moreover some difficulties were faced to reproduce DrLIM, especially on NORB due to the lack of directives for parameters and due to the initial randomness of model initialization which changes the final embedding consequently.
During these experiments, one if not the most important factor for training a Siamese network comes from the pairing strategy.
On this regard, many small factors can have a big impact over the final results such as: the ratio of similar/dissimilar pairs, their ordering (shuffled or not) and the addition/removal of certain pairs.
The new models are easier to train because their convergence to the above results required nearly no effort: the default weights and margin were safely used and the models converged to the expected solutions unlike DrLIM in our experiments.

However there is still room for improvements, especially for the quantitative measures which does not always reflect the structure quality quantified by a visual inspection.
It is important to keep in mind that the visual quality is not necessarily representative of its real usability as features for further layers in a NN.
Nonetheless, it shows that the new model can create quite interesting and human-friendly representation by simple methods.

\chapter{Conclusion}
\label{chap:conclusion}

Neural Networks are widely used in many applications of Computer Vision and their usage is growing every day.
The recent usage of NN in dimensionality reduction is an important step to create better embeddings.
Indeed they learn powerful mappings whose relation seems more deeply tied to the dataset.
In particular they can learn invariance to non-linear distortions while preserving relations between the other distortions.
In the experiments, NN were found to be much faster and reliable for our application than even the state-of-the-art methods such as t-SNE and the only prior-knowledge is the labels or the viewpoint angles.
Moreover, many tricks developed during the past years of researches on NN have helped in this task: CNN architectures for images, Siamese networks and the contrastive loss function to train the new specialized networks.
Thanks to these findings and the current advance in Deep Learning, this work benefited from their improvement in the automatic derivation of better features to represent the data.

In this thesis, the performance of t-SNE on data-augmented datasets were presented with its advantages and shortcomings due to its unsupervised nature.
Another solution was then chosen based on NN because they have many advantages, in particular that they can represent very complex mappings.
To achieve more predictable embeddings, an extension to DrLIM was designed in this work.
Then, several NN models were experimented on different datasets to compare the state-of-the-art with this new solution which demonstrated similar results, while being more general.
In particular, with the new method it is possible to quantify distortions in a consistent way, by just looking at the components of the learned feature representation.
The qualitative results were presented using plots which were analyzed to understand the models.
But quantitative measures were also introduced based on the loss function to compare more objectively with the previous work.
As said, there is still room for improvements in this regard as current measures does not seem completely reliable to measure the quality.
Moreover, a better metric to give a more accurate score was a subject of our research to compare clustering and the predictability of the new models.
However, such a solution is non-trivial to find and this would require comparing the new models with a simple baseline, which, to the best of our knowledge, does not exist at this day.

The current direction of dimensionality reduction started with DrLIM looks very similar to regression problems and this is also the case for the proposed generalization for N-dimensions.
However, regression is more constrained than the current formulation because: (1) in this work, clusters are not constrained to be at a particular position, but only their relative positions matter whereas a regression based approach would set them in absolute positions; (2) the margin formulation allows to put extra gap to make space for clusters without cost whereas regression penalizes in every directions.
Thus, the current formulation allows more freedom in placement than regression and a possibility is that regression would perform worse due to the addition of these artificial constraints but the opposite may be true.
Future work could compare the new models with regression models to see if further restriction would actually help instead.

Another aspect that needs to be explored in future research is to precisely test the predictability of the embeddings.
For instance, the learned features could be used in a larger classification network to predict the distortion intensity and the digit class in MNIST.
Another useful contribution would find a measure that quantifies the ``predictability'' of the new model for certain application.
One simple example would be to use the accuracy of the classification model above but it would not represent the effective usability in practical applications which combine both distortions and application-specific features.
A better experiment is to train a regression model, for instance, to predict the distorted inputs' features given a single input's features.
The accuracy of such a model would serve as a score for predictability.


An application is possible in Biomedical Imaging for classification tasks (\eg: segmentation) which need to detect orientation-dependent structures in images.
The classification of such images begins by a normalization step with a rotation and a translation to normalize the structure of interest into a centered slice.
This slice is used to extract orientation-dependant features which are then fed into a classifier like a CNN.
It is important to understand that the normalization step to extract features is costly especially for 3D images and even specialized FPGAs were designed to compute 3D expensive rotations.
The presented contribution allows theoretically to skip this normalization step to directly create enhanced features.
In this case, the model learns features predictable with respect to rotations and the features for other distortions are easily derived and classified.



\bibliography{thesis}{}

\begin{thebibliography}{10}

\bibitem{bastien2012theano}
Fr{\'e}d{\'e}ric Bastien, Pascal Lamblin, Razvan Pascanu, James Bergstra, Ian
  Goodfellow, Arnaud Bergeron, Nicolas Bouchard, David Warde-Farley, and Yoshua
  Bengio.
\newblock Theano: new features and speed improvements.
\newblock {\em arXiv preprint arXiv:1211.5590}, 2012.

\bibitem{bromley1993signature}
Jane Bromley, James~W Bentz, L{\'e}on Bottou, Isabelle Guyon, Yann LeCun, Cliff
  Moore, Eduard S{\"a}ckinger, and Roopak Shah.
\newblock Signature verification using a “siamese” time delay neural
  network.
\newblock {\em International Journal of Pattern Recognition and Artificial
  Intelligence}, 7(04):669--688, 1993.

\bibitem{chopra2005learning}
Sumit Chopra, Raia Hadsell, and Yann LeCun.
\newblock Learning a similarity metric discriminatively, with application to
  face verification.
\newblock In {\em Computer Vision and Pattern Recognition, 2005. CVPR 2005.
  IEEE Computer Society Conference on}, volume~1, pages 539--546. IEEE, 2005.

\bibitem{ciresan2011flexible}
Dan~C Ciresan, Ueli Meier, Jonathan Masci, Luca Maria~Gambardella, and
  J{\"u}rgen Schmidhuber.
\newblock Flexible, high performance convolutional neural networks for image
  classification.
\newblock In {\em IJCAI Proceedings-International Joint Conference on
  Artificial Intelligence}, volume~22, page 1237, 2011.

\bibitem{coates2013deep}
Adam Coates, Brody Huval, Tao Wang, David Wu, Bryan Catanzaro, and Ng~Andrew.
\newblock Deep learning with cots hpc systems.
\newblock In {\em Proceedings of the 30th international conference on machine
  learning}, pages 1337--1345, 2013.

\bibitem{collobert2011torch7}
Ronan Collobert, Koray Kavukcuoglu, and Cl{\'e}ment Farabet.
\newblock Torch7: A matlab-like environment for machine learning.
\newblock In {\em BigLearn, NIPS Workshop}, number EPFL-CONF-192376, 2011.

\bibitem{cox2000multidimensional}
Trevor~F Cox and Michael~AA Cox.
\newblock {\em Multidimensional scaling}.
\newblock CRC Press, 2000.

\bibitem{csaji2001approximation}
Bal{\'a}zs~Csan{\'a}d Cs{\'a}ji.
\newblock Approximation with artificial neural networks.
\newblock {\em Faculty of Sciences, Etvs Lornd University, Hungary}, 24, 2001.

\bibitem{dai2014document}
Andrew~M Dai, Christopher Olah, Quoc~V Le, and Greg~S Corrado.
\newblock Document embedding with paragraph vectors.
\newblock In {\em NIPS Deep Learning Workshop}, 2014.

\bibitem{donahue2013decaf}
Jeff Donahue, Yangqing Jia, Oriol Vinyals, Judy Hoffman, Ning Zhang, Eric
  Tzeng, and Trevor Darrell.
\newblock Decaf: A deep convolutional activation feature for generic visual
  recognition.
\newblock {\em arXiv preprint arXiv:1310.1531}, 2013.

\bibitem{gens2014deep}
Robert Gens and Pedro~M Domingos.
\newblock Deep symmetry networks.
\newblock In {\em Advances in neural information processing systems}, pages
  2537--2545, 2014.

\bibitem{goodfellow2009measuring}
Ian Goodfellow, Honglak Lee, Quoc~V Le, Andrew Saxe, and Andrew~Y Ng.
\newblock Measuring invariances in deep networks.
\newblock In {\em Advances in neural information processing systems}, pages
  646--654, 2009.

\bibitem{goodfellow2013pylearn2}
Ian~J Goodfellow, David Warde-Farley, Pascal Lamblin, Vincent Dumoulin, Mehdi
  Mirza, Razvan Pascanu, James Bergstra, Fr{\'e}d{\'e}ric Bastien, and Yoshua
  Bengio.
\newblock Pylearn2: a machine learning research library.
\newblock {\em arXiv preprint arXiv:1308.4214}, 2013.

\bibitem{hadsell2006dimensionality}
Raia Hadsell, Sumit Chopra, and Yann LeCun.
\newblock Dimensionality reduction by learning an invariant mapping.
\newblock In {\em Computer vision and pattern recognition, 2006 IEEE computer
  society conference on}, volume~2, pages 1735--1742. IEEE, 2006.

\bibitem{SNE}
Geoffrey~E Hinton and Sam~T Roweis.
\newblock Stochastic neighbor embedding.
\newblock In {\em Advances in neural information processing systems}, pages
  833--840, 2002.

\bibitem{jia2014caffe}
Yangqing Jia, Evan Shelhamer, Jeff Donahue, Sergey Karayev, Jonathan Long, Ross
  Girshick, Sergio Guadarrama, and Trevor Darrell.
\newblock Caffe: Convolutional architecture for fast feature embedding.
\newblock In {\em Proceedings of the ACM International Conference on
  Multimedia}, pages 675--678. ACM, 2014.

\bibitem{krizhevsky2012imagenet}
Alex Krizhevsky, Ilya Sutskever, and Geoffrey~E Hinton.
\newblock Imagenet classification with deep convolutional neural networks.
\newblock In {\em Advances in neural information processing systems}, pages
  1097--1105, 2012.

\bibitem{lawrence1997face}
Steve Lawrence, C~Lee Giles, Ah~Chung Tsoi, and Andrew~D Back.
\newblock Face recognition: A convolutional neural-network approach.
\newblock {\em Neural Networks, IEEE Transactions on}, 8(1):98--113, 1997.

\bibitem{lecun1998mnist}
Yann LeCun and Corinna Cortes.
\newblock The mnist database of handwritten digits, 1998.

\bibitem{lecun2004learning}
Yann LeCun, Fu~Jie Huang, and Leon Bottou.
\newblock Learning methods for generic object recognition with invariance to
  pose and lighting.
\newblock In {\em Computer Vision and Pattern Recognition, 2004. CVPR 2004.
  Proceedings of the 2004 IEEE Computer Society Conference on}, volume~2, pages
  II--97. IEEE, 2004.

\bibitem{nair2010rectified}
Vinod Nair and Geoffrey~E Hinton.
\newblock Rectified linear units improve restricted boltzmann machines.
\newblock In {\em Proceedings of the 27th International Conference on Machine
  Learning (ICML-10)}, pages 807--814, 2010.

\bibitem{nasse2009face}
Fabian Nasse, Christian Thurau, and Gernot Fink.
\newblock Face detection using gpu-based convolutional neural networks.
\newblock In {\em Computer Analysis of Images and Patterns}, pages 83--90.
  Springer, 2009.

\bibitem{pedregosa2011scikit}
Fabian Pedregosa, Ga{\"e}l Varoquaux, Alexandre Gramfort, Vincent Michel,
  Bertrand Thirion, Olivier Grisel, Mathieu Blondel, Peter Prettenhofer, Ron
  Weiss, Vincent Dubourg, et~al.
\newblock Scikit-learn: Machine learning in python.
\newblock {\em The Journal of Machine Learning Research}, 12:2825--2830, 2011.

\bibitem{prechelt1994proben1}
Lutz Prechelt et~al.
\newblock Proben1: A set of neural network benchmark problems and benchmarking
  rules.
\newblock 1994.

\bibitem{roweis2000nonlinear}
Sam~T Roweis and Lawrence~K Saul.
\newblock Nonlinear dimensionality reduction by locally linear embedding.
\newblock {\em Science}, 290(5500):2323--2326, 2000.

\bibitem{rowley1998neural}
Henry~A Rowley, Shumeet Baluja, and Takeo Kanade.
\newblock Neural network-based face detection.
\newblock {\em Pattern Analysis and Machine Intelligence, IEEE Transactions
  on}, 20(1):23--38, 1998.

\bibitem{schmidhuber2015deep}
J{\"u}rgen Schmidhuber.
\newblock Deep learning in neural networks: An overview.
\newblock {\em Neural Networks}, 61:85--117, 2015.

\bibitem{simard2003best}
Patrice~Y Simard, Dave Steinkraus, and John~C Platt.
\newblock Best practices for convolutional neural networks applied to visual
  document analysis.
\newblock In {\em 2013 12th International Conference on Document Analysis and
  Recognition}, volume~2, pages 958--958. IEEE Computer Society, 2003.

\bibitem{szegedy2013intriguing}
Christian Szegedy, Wojciech Zaremba, Ilya Sutskever, Joan Bruna, Dumitru Erhan,
  Ian Goodfellow, and Rob Fergus.
\newblock Intriguing properties of neural networks.
\newblock {\em arXiv preprint arXiv:1312.6199}, 2013.

\bibitem{taylor2011learning}
Graham~W Taylor, Ian Spiro, Christoph Bregler, and Rob Fergus.
\newblock Learning invariance through imitation.
\newblock In {\em Computer Vision and Pattern Recognition (CVPR), 2011 IEEE
  Conference on}, pages 2729--2736. IEEE, 2011.

\bibitem{tenenbaum2000global}
Joshua~B Tenenbaum, Vin De~Silva, and John~C Langford.
\newblock A global geometric framework for nonlinear dimensionality reduction.
\newblock {\em Science}, 290(5500):2319--2323, 2000.

\bibitem{van2009new}
Laurens Van~der Maaten.
\newblock A new benchmark dataset for handwritten character recognition.
\newblock {\em Tilburg University}, pages 2--5, 2009.

\bibitem{van2013barnes}
Laurens van~der Maaten.
\newblock Barnes-hut-sne.
\newblock {\em arXiv preprint arXiv:1301.3342}, 2013.

\bibitem{t-SNE}
Laurens Van~der Maaten and Geoffrey Hinton.
\newblock Visualizing data using t-sne.
\newblock {\em Journal of Machine Learning Research}, 9(2579-2605):85, 2008.

\bibitem{mnist_web}
Christopher J.C.~Burges Yann~LeCun, Corinna~Cortes.
\newblock {The MNIST Database of handwritten digits}.
\newblock \url{http://yann.lecun.com/exdb/mnist/}.
\newblock [Online; accessed 01-June-2015].

\bibitem{yaotiny}
Leon Yao and John Miller.
\newblock Tiny imagenet classification with convolutional neural networks.

\bibitem{yu2014visualizing}
Wei Yu, Kuiyuan Yang, Yalong Bai, Hongxun Yao, and Yong Rui.
\newblock Visualizing and comparing convolutional neural networks.
\newblock {\em arXiv preprint arXiv:1412.6631}, 2014.

\end{thebibliography}
\bibliographystyle{plain}

\nocite{lecun2004learning}

\end{document}